\documentclass{article} 
\usepackage{iclr2027_conference,times}

\usepackage{amsmath,amsfonts,bm}

\def\eqref#1{equation~\ref{#1}}

\def\1{\bm{1}}

\DeclareMathAlphabet{\mathsfit}{\encodingdefault}{\sfdefault}{m}{sl}
\SetMathAlphabet{\mathsfit}{bold}{\encodingdefault}{\sfdefault}{bx}{n}

\usepackage{hyperref}
\usepackage{url}

\usepackage{microtype}
\usepackage{graphicx}
\usepackage{subcaption}
\usepackage{booktabs} 
\usepackage[utf8]{inputenc}
\usepackage{tabularx} 
\usepackage{makecell} 

\usepackage{amsmath}
\usepackage{amssymb}
\usepackage{mathtools}
\usepackage{amsthm}
\usepackage{multirow}
\usepackage[table,xcdraw]{xcolor}
\usepackage[normalem]{ulem}
\useunder{\uline}{\ul}{}

\usepackage{wasysym}
\usepackage{adjustbox}
\usepackage{algorithm}
\usepackage{algorithmic}
\usepackage{wrapfig}

\title{BITS: Rethinking Fair and Comprehensive Evaluation for Irregular Time Series Forecasting}

\author{%
\textbf{Kangjia Yan}\textsuperscript{1}\enspace
\textbf{Linfeng Wang}\textsuperscript{1}\enspace
\textbf{Tianen Shen}\textsuperscript{1}\enspace
\textbf{Xiangfei Qiu}\textsuperscript{1}\enspace
\textbf{Ruitong Zhang}\textsuperscript{1}\\[0.5em]
\textbf{Hao Miao}\textsuperscript{2}\enspace
\textbf{Jilin Hu}\textsuperscript{1}\enspace
\textbf{Chenjuan Guo}\textsuperscript{1}\enspace
\textbf{Bin Yang}\textsuperscript{1}\enspace
\textbf{Christian S. Jensen}\textsuperscript{3}%
}

\iclrfinalcopy 
\begin{document}

\maketitle
\lhead{Preprint}

\maketitle
\lhead{Preprint}

\maketitle
\lhead{Preprint}

\begingroup
\renewcommand{\thefootnote}{}
\footnotetext{%
\raggedright
\noindent
\textsuperscript{1}East China Normal University,
Shanghai, China.
\textsuperscript{2}University of Electronic Science
and Technology of China, Chengdu, China.
\textsuperscript{3}Aalborg University, Aalborg, Denmark.
Emails (in author order):
\texttt{kjyan@stu.ecnu.edu.cn},
\texttt{lfwang@stu.ecnu.edu.cn},
\texttt{teshen@stu.ecnu.edu.cn},
\texttt{xfqiu@stu.ecnu.edu.cn},
\texttt{rtzhang@stu.ecnu.edu.cn},
\texttt{hao-miao@outlook.com},
\texttt{jlhu@dase.ecnu.edu.cn},
\texttt{cjguo@dase.ecnu.edu.cn},
\texttt{byang@dase.ecnu.edu.cn},
\texttt{csj@cs.aau.dk}.
\par
}
\endgroup

\begin{abstract}

Despite recent progress in irregular time series forecasting, the field still lacks a unified benchmark for fair and comprehensive evaluation. Existing evaluations are often conducted on a limited set of datasets with inconsistent experimental protocols and predominantly error-based metrics, rendering it difficult to compare and assess methods fairly and comprehensively across diverse settings. To eliminate these limitations and accelerate progress, we propose BITS, a standardized, reproducible, and extensible benchmark for advancing research on irregular time series forecasting. BITS covers eleven datasets from nine domains with diverse irregularity characteristics, and it characterizes the datasets according to their missing rate, missing pattern complexity, sampling irregularity, and skewness. Further, it offers a unified pipeline for data preprocessing, model integration and evaluation, and reporting. It accommodates regular and irregular time series forecasting methods, including time series foundation models, under consistent settings, incorporating both error-based and non-error-based evaluation metrics. Findings include that method performance varies substantially across irregularity characteristics, with no single modeling strategy consistently dominating. We also find that using error-based or non-error-based metrics can yield different model rankings, highlighting the need for multi-dimensional evaluation. The code can be found at \url{https://anonymous.4open.science/r/BITS-8F2E/}.
\end{abstract}

\section{Introduction}
\label{introsection}


Irregular time series are widely encountered in real-world scenarios, such as healthcare monitoring and environmental sensing. In recent years, a variety of methods have been proposed for irregular time series forecasting, such as temporal segmentation and aggregation methods~\citep{DBLP:conf/icml/LuoZ0025, DBLP:conf/icml/ZhangYL0024, qiu2026bridging, DBLP:conf/aaai/ZhouHWWKL26}, differential-equation-based models~\citep{DBLP:conf/nips/MercataliFC24} and data-format transformation methods~\citep{DBLP:conf/aaai/YalavarthiMSABJ24, DBLP:conf/icml/LiL0ZL025, liu2026astgi}. These methods have advanced the modeling of irregular time series and achieved promising performance in irregular time series forecasting. However, despite this progress, the community still lacks a systematic, standardized, and representative benchmark for irregular time series forecasting, making it difficult to fairly compare different methods and comprehensively evaluate their effectiveness under diverse irregular settings. We identify two key limitations in existing evaluations of irregular time series forecasting.

\begin{figure}[t]
    \centering
    \includegraphics[width=\linewidth]{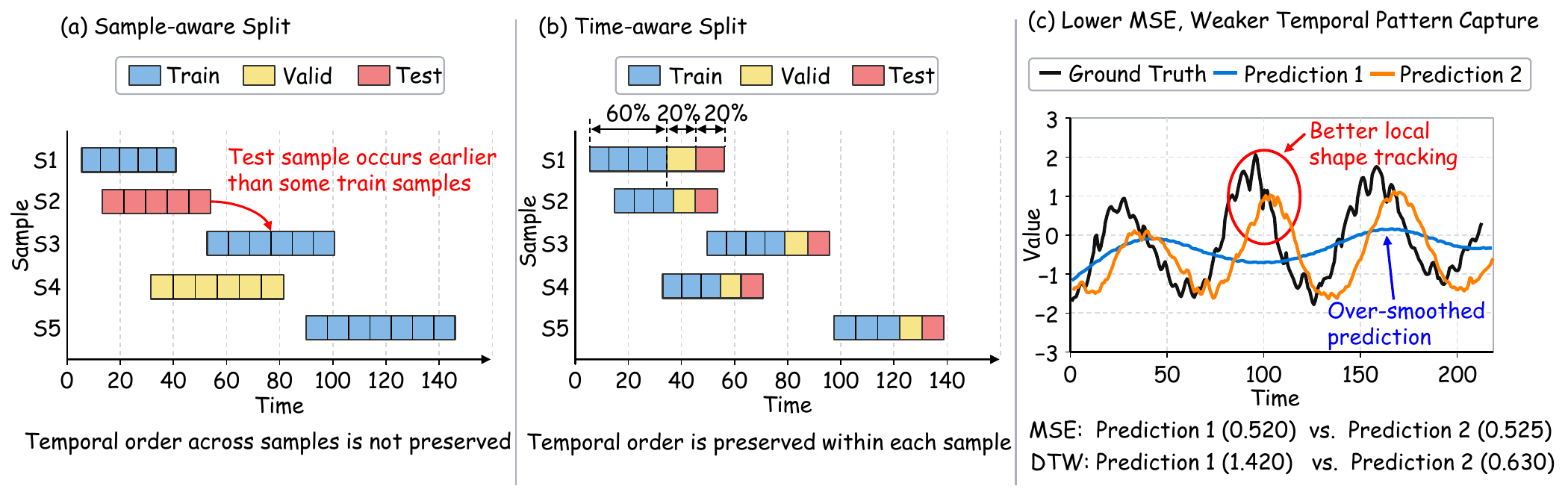}
    \caption{(a) Sample-aware splitting strategy. (b) Time-aware splitting strategy. (c) The limitation of error-based metrics.}
    \label{fig:intro}
    \vspace{-0.5cm}
\end{figure}

\textbf{First, existing studies provide limited dataset coverage and splitting strategies.} (i) Many existing studies only rely on a few datasets such as PhysioNet, MIMIC, Human Activity and USHCN~\citep{DBLP:conf/icml/LuoZ0025, DBLP:conf/icml/ZhangYL0024, DBLP:conf/icml/LiL0ZL025}, which cover a relatively narrow range of application domains and irregularity characteristics. In addition, some datasets exhibit extremely high missing rates and complex missing patterns~\citep{silva2012predicting}, failing to fully reflect the missing mechanisms commonly observed in real-world environments. (ii) Moreover, many studies adopt a sample-aware splitting strategy without considering temporal order~\citep{DBLP:conf/aaai/ZhouHWWKL26, DBLP:conf/aaai/YalavarthiMSABJ24}, as shown in Figure~\ref{fig:intro} (a). Although such splitting can be necessary for sparsely observed datasets, where temporal splitting may leave insufficient observations for meaningful forecasting windows, it is also widely used for datasets with a sufficient number of observations. In these cases, ignoring temporal order may allow future observations to appear in the training set while earlier observations are used for validation or testing, potentially introducing information leakage.

\textbf{Second, existing evaluation protocols are neither standardized nor comprehensive.} (i) Different studies adopt different normalization, making reported performance dependent not only on model capability but also on experimental configuration~\citep{DBLP:conf/icml/ZhangYL0024, DBLP:conf/kdd/ZhangZCBL23}. In particular, different normalization strategies can lead to different data scales for the same dataset, making regression metrics such as MSE difficult to compare fairly across methods. (ii) Existing studies also employ inconsistent input-output window settings, with different combinations of lookback and prediction lengths across datasets and methods~\citep{DBLP:conf/icml/LiL0ZL025, DBLP:journals/corr/abs-2505-11250}. Some settings, such as using only 3 hours of historical observations to forecast the following 45 hours~\citep{DBLP:conf/icml/ZhangYL0024}, may be difficult to interpret in practical forecasting scenarios. (iii) In addition, most evaluations mainly rely on error-based metrics such as MSE~\citep{DBLP:conf/iclr/ShuklaM21, DBLP:conf/nips/BilosSRJG21, DBLP:conf/icml/SchirmerELR22}, which measure point-wise prediction errors and may fail to reflect whether a model captures the temporal dynamics. For example, a model may achieve relatively low MSE while producing nearly constant predictions across the forecast horizon, as shown in Figure~\ref{fig:intro} (c). Such predictions may obtain favorable regression errors despite failing to capture important temporal patterns. Therefore, low point-wise errors alone do not necessarily indicate effective forecasting of temporal dynamics. 


    

This study addresses the above challenges by providing BITS, a fair and comprehensive \textbf{B}enchmark for \textbf{I}rregular \textbf{T}ime \textbf{S}eries forecasting. BITS is designed to support fair, reproducible, and extensible evaluation across datasets, models, and metrics, as shown in Table~\ref{tab:benchmark_comparison}. \textbf{To address the first limitation, BITS expands the dataset coverage and splitting strategy.} (i) We collect eleven datasets from nine domains with diverse irregularity characteristics (e.g., missing rate), providing broader coverage of real-world irregular scenarios. (ii) We adopt time-aware splitting strategies for datasets with sufficient observations for temporal splitting to preserve temporal order, as shown in Figure~\ref{fig:intro} (b), while sample-aware splitting is retained for sparsely observed datasets. Under time-aware dataset splitting strategies, each sample is divided according to temporal order and predefined proportions. The corresponding segments from all instances are then merged to form the final training, validation and test sets, reducing the risk of temporal information leakage. 

\textbf{To address the second limitation, BITS establishes a standardized and comprehensive evaluation pipeline.} (i) We standardize key procedures, including normalization, model integration, evaluation, and reporting, and apply consistent normalization strategies (e.g., min-max) to reduce scale discrepancies and enable fair comparisons across methods. (ii) We define customized lookback and prediction lengths for each dataset, which are selected according to the temporal characteristics and practical forecasting requirements of each dataset. (iii) Beyond error-based metrics, BITS also incorporates non-error-based metrics to evaluate trend consistency, shape similarity and structural changes, which enables a more comprehensive understanding of whether a model captures the temporal patterns rather than merely achieving low point-wise errors. In addition, BITS supports irregular forecasting models, regular forecasting models and time series foundation models under the same benchmarking framework, allowing systematic comparison across different modeling paradigms.


In summary, our main contributions are as follows:

\begin{itemize}
    \item We propose BITS, a comprehensive benchmark for irregular time series forecasting. BITS supports fair and reproducible evaluation of specialized irregular forecasting models, regular forecasting models and time series foundation models across diverse irregular datasets.
    \item We construct a diverse dataset collection with systematic irregularity analysis. BITS includes eleven datasets from nine domains and characterizes them from multiple dimensions, including missing rate, missing pattern complexity, sampling irregularity, and skewness.
    \item We design a unified and extensible benchmarking pipeline. BITS standardizes preprocessing, normalization, dataset splitting, window construction, model integration, evaluation, and reporting, reducing experimental bias and improving reproducibility. In addition to error-based metrics, BITS also incorporates non-error-based metrics to evaluate whether models capture temporal dynamics.
    \item We conduct extensive experiments and derive empirical insights into irregular forecasting. The results reveal substantial performance variation across irregularity characteristics and modeling paradigms, while different evaluation metrics can lead to different assessments of model performance.
\end{itemize}

\begin{table}[t]
\centering
\caption{Comparison of BITS with existing time series forecasting benchmarks.
$\checkmark$ indicates present, $\times$ indicates absent, and $\ocircle$ indicates incomplete.}
\label{tab:benchmark_comparison}

\footnotesize
\setlength{\tabcolsep}{1.3pt}
\renewcommand{\arraystretch}{1.05}

\begin{adjustbox}{max width=\linewidth}
\begin{tabular}{@{}clccccccccc@{}}
\toprule
& \textbf{Benchmark} & 
\makecell{Multi\\Domain} & 
\makecell{Irregularity\\Analysis} & 
\makecell{Irregular\\TS} & 
\makecell{Irregular\\Forecasting} & 
\makecell{Regular\\Model} & 
\makecell{Irregular\\Model} & 
\makecell{TS Foundation\\Model} & 
\makecell{Scalable\\Pipeline} & 
\makecell{Non-error-based\\Metric} \\
\midrule

\multirow{6}{*}{\rotatebox[origin=c]{90}{\textbf{Regular}}}
& BasicTS~\citep{liang2022basicts}     
& $\checkmark$ & $\times$ & $\times$ 
& $\times$ & $\checkmark$ & $\times$ 
& $\times$ & $\ocircle$ & $\times$ \\

& BasicTS+~\citep{DBLP:journals/tkde/ShaoWXWYZYSJCCJC25}    
& $\checkmark$ & $\times$ & $\times$ 
& $\times$ & $\checkmark$ & $\times$ 
& $\times$ & $\ocircle$ & $\times$ \\

& Monash~\citep{DBLP:conf/nips/GodahewaBWHM21}      
& $\checkmark$ & $\times$ & $\times$ 
& $\times$ & $\checkmark$ & $\times$ 
& $\times$ & $\ocircle$ & $\times$ \\

& TSLib~\citep{wang2026deep}       
& $\checkmark$ & $\times$ & $\times$ 
& $\times$ & $\checkmark$ & $\times$ 
& $\times$ & $\ocircle$ & $\times$ \\

& TFB~\citep{qiu2024tfb}         
& $\checkmark$ & $\times$ & $\times$ 
& $\times$ & $\checkmark$ & $\times$ 
& $\times$ & $\checkmark$ & $\times$ \\

& TSFM-Bench~\citep{li2025tsfm}  
& $\checkmark$ & $\times$ & $\times$ 
& $\times$ & $\checkmark$ & $\times$ 
& $\checkmark$ & $\checkmark$ & $\times$ \\

\midrule

\multirow{6}{*}{\rotatebox[origin=c]{90}{\textbf{Irregular}}}
& MIMIC-III~\citep{Harutyunyan2019}   
& $\times$ & $\times$ & $\checkmark$ 
& $\times$ & $\checkmark$ & $\times$ 
& $\times$ & $\ocircle$ & $\times$ \\

& MIMIC-IV~\citep{bui2024benchmarking}    
& $\times$ & $\times$ & $\checkmark$ 
& $\times$ & $\checkmark$ & $\times$ 
& $\times$ & $\ocircle$ & $\times$ \\

& HyperIMTS~\citep{DBLP:conf/icml/LiL0ZL025}   
& $\ocircle$ & $\times$ & $\checkmark$ 
& $\checkmark$ & $\checkmark$ & $\checkmark$ 
& $\times$ & $\ocircle$ & $\times$ \\

& tPatchGNN~\citep{DBLP:conf/icml/ZhangYL0024}  
& $\ocircle$ & $\ocircle$ & $\checkmark$ 
& $\checkmark$ & $\checkmark$ & $\checkmark$ 
& $\times$ & $\ocircle$ & $\times$ \\

& Time-IMM~\citep{DBLP:conf/nips/ChangHSWWPC25}    
& $\checkmark$ & $\checkmark$ & $\checkmark$ 
& $\checkmark$ & $\checkmark$ & $\checkmark$ 
& $\times$ & $\ocircle$ & $\times$ \\

& \textbf{BITS (Ours)} 
& $\checkmark$ & $\checkmark$ & $\checkmark$ 
& $\checkmark$ & $\checkmark$ & $\checkmark$ 
& $\checkmark$ & $\checkmark$ & $\checkmark$ \\

\bottomrule
\end{tabular}
\end{adjustbox}

\end{table}


\section{Related Work}
\label{sec:related work}

\textbf{Irregular Time Series Forecasting.} Existing irregular time series forecasting methods adopt diverse strategies to handle irregular observations. Imputation-based methods reconstruct or aggregate sparse observations into regular representations, such as mTAN~\citep{DBLP:conf/iclr/ShuklaM21}. Temporal segmentation and aggregation methods, such as tPatchGNN~\citep{DBLP:conf/icml/ZhangYL0024}, Hi-Patch~\citep{DBLP:conf/icml/LuoZ0025} and TiWeaver~\citep{DBLP:conf/kdd/LiTMLGY26}, organize irregular observations into adaptive temporal units for representation learning. Multi-scale methods such as Warpformer~\citep{DBLP:conf/kdd/ZhangZCBL23} capture temporal patterns at different resolutions. Continuous-time approaches, such as NeuralFlows~\citep{DBLP:conf/nips/BilosSRJG21}, CRU~\citep{DBLP:conf/icml/SchirmerELR22}, and GNeuralFlow~\citep{DBLP:conf/nips/MercataliFC24}, model dynamics directly in continuous time. Other methods transform observations into structured representations such as sets or graphs, including SeFT~\citep{DBLP:conf/icml/HornMBRB20}, GraFITi~\citep{DBLP:conf/aaai/YalavarthiMSABJ24}, HyperIMTS~\citep{DBLP:conf/icml/LiL0ZL025}, or explicitly incorporate sampling context, as in GRU-D~\citep{DBLP:journals/corr/ChePCSL16}.

\textbf{Benchmarks for Time Series Forecasting.} Existing time series benchmarks mainly focus on regularly sampled forecasting. BasicTS~\citep{liang2022basicts}, BasicTS+~\citep{DBLP:journals/tkde/ShaoWXWYZYSJCCJC25}, Monash~\citep{DBLP:conf/nips/GodahewaBWHM21}, TSLib~\citep{wang2026deep}, and TFB~\citep{qiu2024tfb} provide standardized datasets and evaluation pipelines for regular forecasting. TSFM-Bench~\citep{li2025tsfm} further extends evaluation to time series foundation models. However, these benchmarks provide limited support for irregular time series forecasting. Several studies have explored evaluation on irregular time series. MIMIC-III~\citep{Harutyunyan2019} and MIMIC-IV~\citep{bui2024benchmarking} primarily focus on clinical applications. HyperIMTS~\citep{DBLP:conf/icml/LiL0ZL025} and tPatchGNN~\citep{DBLP:conf/icml/ZhangYL0024} evaluate irregular forecasting models on a limited number of datasets. More recently, Time-IMM~\citep{DBLP:conf/nips/ChangHSWWPC25} extends irregular evaluation toward multimodal representation learning. As summarized in Table~\ref{tab:benchmark_comparison}, existing benchmarks still lack a unified framework that jointly covers diverse irregular datasets, irregularity analysis, regular and irregular forecasting models, TSFMs, scalable evaluation pipelines, and multi-dimensional metrics. BITS is designed to bridge these gaps through systematic and unified evaluation of irregular time series forecasting.

\section{BITS Benchmark Framework}
\label{sec: ITFB Benchmark Framework}

\begin{figure}[t]
    \centering
    \includegraphics[width=\linewidth]{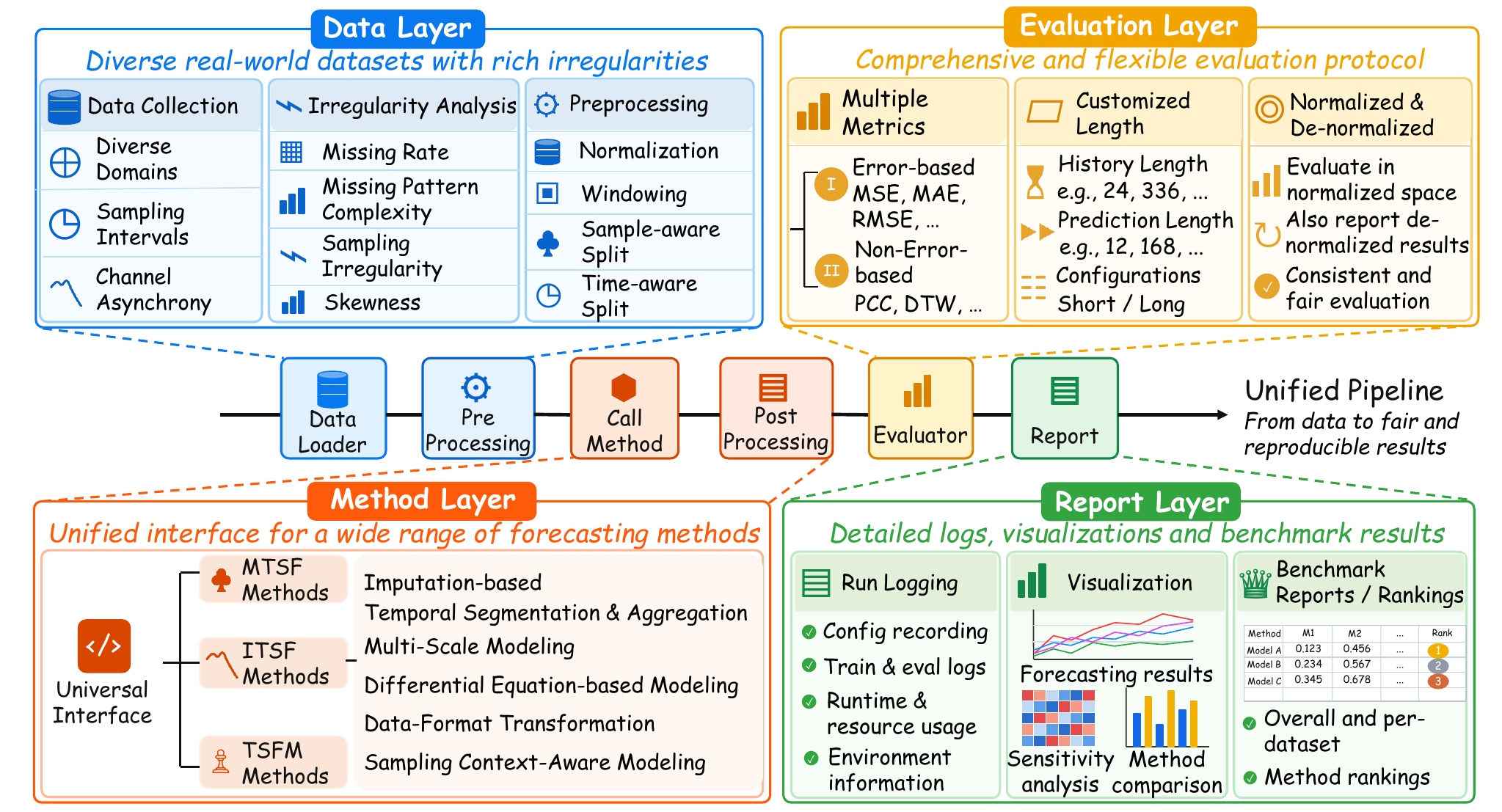}
    \caption{Overview of the BITS pipeline.}
    \label{fig:ITFB}
    \vspace{-0.5cm}
\end{figure}

\vspace{-0.2cm}
\subsection{Datasets Details}

We collect eleven datasets spanning nine domains, such as activity and climate. For datasets containing samples with substantially different durations or numbers of observations, we select a subset of samples and variables so that the retained samples have comparable temporal spans and observation counts, 
\begin{wrapfigure}{r}{0.60\textwidth}
    \centering
    \includegraphics[width=\linewidth]{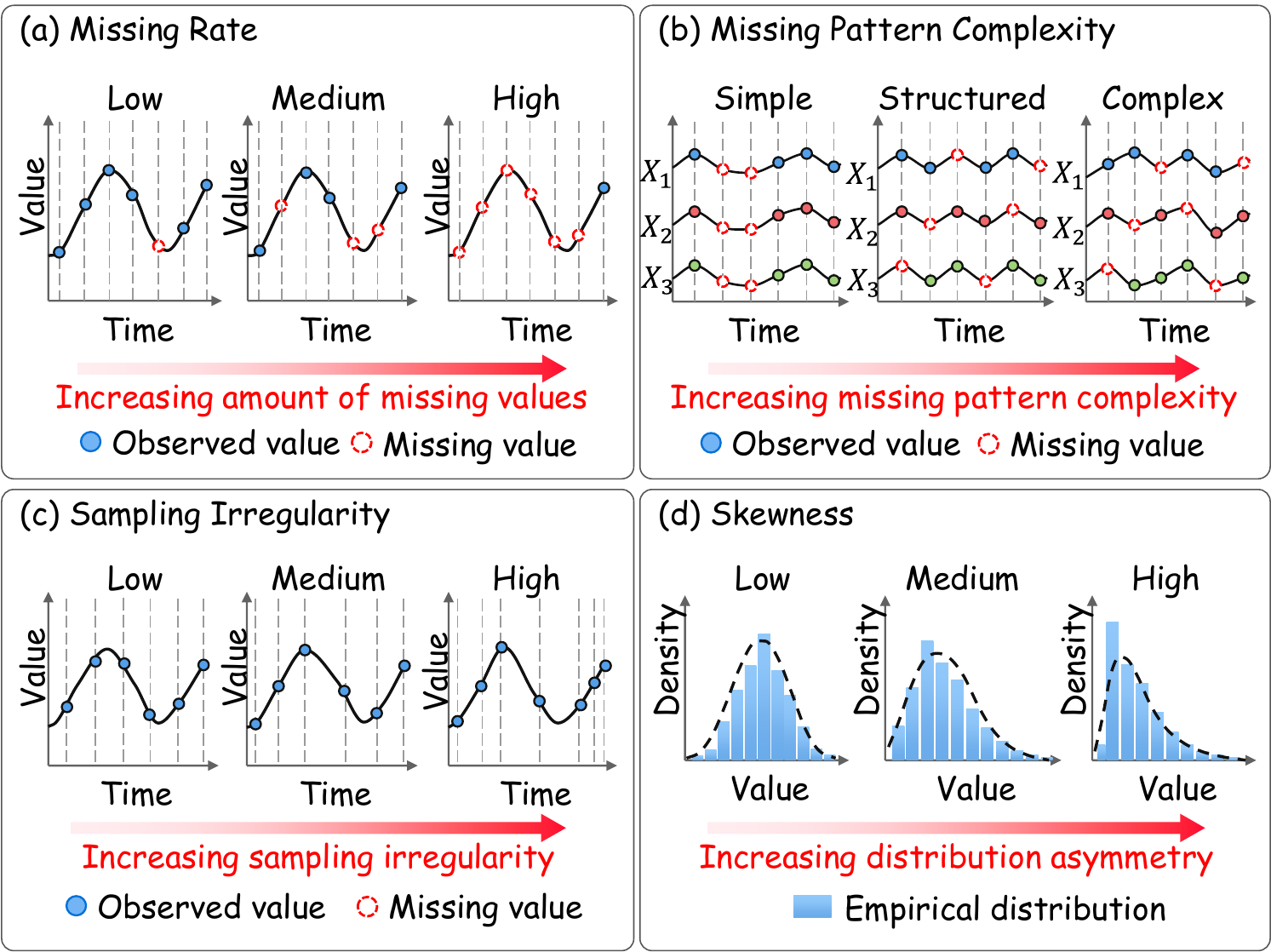}
    \caption{Illustration of four irregularity characteristics.}
    \label{fig:irregularcharacteristic}
    \vspace{-0.5cm}
\end{wrapfigure}
facilitating consistent window construction across samples. Then we process them into a unified standard format. Detailed descriptions and statistical information about the datasets can be found in Appendix~\ref{datasets_appendix}. We adopt different splitting strategies according to the observation counts of each dataset. For sparsely observed datasets (e.g., PhysioNet, USHCN, and APTC), we adopt sample-aware splitting which randomly divides all samples in each dataset, because time-aware splitting would leave insufficient observations for meaningful forecasting window construction. For the remaining datasets which provide sufficient observations, we adopt time-aware splitting by dividing each sample chronologically and merging the corresponding segments across samples. This overcomes the shortcoming of existing irregular time series forecasting methods that do not consider the time order in dataset splitting even when the observation counts of the dataset are sufficient to support time-aware splitting. 

We characterize dataset irregularity from four dimensions, including missing rate, missing pattern complexity, sampling irregularity and skewness, as shown in Figure~\ref{fig:irregularcharacteristic}.
Detailed definitions and formulas are provided in Appendix~\ref{data_characteristics_appendix}. To further illustrate dataset diversity, we visualize the distributions of the four irregularity characteristics in a shared PCA space in Figure~\ref{fig:coverage} in 
Appendix~\ref{data_characteristics_appendix}. Compared with commonly used datasets such as PhysioNet, Human Activity, and USHCN, BITS covers a broader region of the characteristic space, highlighting the rich diversity of irregularity characteristics across its eleven datasets.

\vspace{-0.2cm}
\subsection{Compared Baselines}
To investigate the advantages and limitations of different methods, we evaluate six categories of irregular time series forecasting methods, including imputation-based, temporal segmentation and aggregation, multi-scale modeling, differential equation-based modeling, data-format transformation and sampling context-aware modeling methods.
For imputation-based methods, we choose mTAN~\citep{DBLP:conf/iclr/ShuklaM21}. In terms of temporal segmentation and aggregation methods, we include tPatchGNN~\citep{DBLP:conf/icml/ZhangYL0024}, Hi-Patch~\citep{DBLP:conf/icml/LuoZ0025}, APN~\citep{DBLP:journals/corr/abs-2505-11250}, TFMixer~\citep{qiu2026bridging}, KAFNet~\citep{DBLP:conf/aaai/ZhouHWWKL26} and TiWeaver~\citep{DBLP:conf/kdd/LiTMLGY26}. For multi-scale modeling methods, we include Warpformer~\citep{DBLP:conf/kdd/ZhangZCBL23}. For differential equation-based modeling methods, we choose NeuralFlows~\citep{DBLP:conf/nips/BilosSRJG21}, GNeuralFlow~\citep{DBLP:conf/nips/MercataliFC24} and CRU~\citep{DBLP:conf/icml/SchirmerELR22}. In terms of data-format transformation methods, we include GraFITi~\citep{DBLP:conf/aaai/YalavarthiMSABJ24}, SeFT~\citep{DBLP:conf/icml/HornMBRB20}, HyperIMTS~\citep{DBLP:conf/icml/LiL0ZL025} and ASTGI~\citep{liu2026astgi}. For sampling context-aware modeling methods, we choose GRU-D~\citep{DBLP:journals/corr/ChePCSL16}. For a more comprehensive and fair comparison, we also include multivariate time series forecasting methods: DLinear~\citep{DBLP:conf/aaai/ZengCZ023}, TimesNet~\citep{wu2023timesnet},
iTransformer~\citep{DBLP:conf/iclr/LiuHZWWML24}, PatchTST~\citep{DBLP:conf/iclr/NieNSK23}, Crossformer~\citep{DBLP:conf/iclr/ZhangY23}, DUET~\citep{DBLP:conf/kdd/QiuW0GH025}, SRSNet~\citep{DBLP:conf/nips/WuQCLHGY25}, and time series foundation models: MIRA~\citep{DBLP:conf/nips/LiDXFSHSSYYB25}, Sundial~\citep{DBLP:conf/icml/LiuQSCY00L25} and Timer~\citep{liu2024timer}.

\vspace{-0.2cm}
\subsection{Evaluation metrics}
Existing methods primarily rely on error-based metrics such as MSE and MAE, and report prediction performance based on normalized data. However, the error-based metrics can be heavily influenced by the data range, making them less intuitive for interpretation. Also, assessing prediction performance on normalized data may result in low error values, which may obscure the practical interpretation of forecasting errors. Furthermore, these metrics are based on distance measures, comparing the predicted values with the ground truth on a per-time-step basis. In scenarios where the model outputs near-constant predictions for the entire forecast horizon, the error-based metrics may fail to provide a meaningful assessment of model performance, as shown in Figure~\ref{fig:intro} (c). Therefore, the evaluation metrics should go beyond simple point-wise distance measures and also consider pattern alignment and temporal structure. 

In order to conduct an effective, comprehensive, and consistent evaluation of the prediction results, we evaluate on both normalized and denormalized data and incorporate additional metrics, including Mean Absolute Percentage Error (MAPE), Symmetric Mean Absolute Percentage Error (SMAPE), Root Mean Squared Error (RMSE), Weighted Absolute Percent Error (WAPE), and Modified Symmetric  Mean Absolute Percentage Error (MSMAPE). In addition, we propose incorporating multi-dimensional metrics that evaluate both value accuracy and pattern similarity, including Pearson Correlation (PCC), Coefficient of Determination ($R^2$), Directional Accuracy (DA), Dynamic Time Warping (DTW), Change Point Number Difference (CPN) and Information Entropy Ratio (IER). This allows for a more holistic assessment of a model's ability to learn and forecast complex temporal structures inherent in irregular time series data. The formulations and detailed descriptions of these metrics are provided in Appendix~\ref{appendix_metrics}. 

\vspace{-0.2cm}
\subsection{Unified Pipeline}
Different data preprocessing procedures and evaluation criteria lead to differences in model performance. To ensure fair and consistent evaluation, we designed a unified pipeline comprising a data layer, a method layer, an evaluation layer, and a reporting layer. A detailed description of each layer is provided below.

\textit{Data Layer} stores comprehensive irregular time series from different domains, encompassing diverse sampling intervals, channel asynchrony and irregularity characteristics. For data preprocessing, we apply a normalization procedure, with min-max normalization as the default choice. The loader selects different dataset splitting strategies, including sample-aware or time-aware splitting, according to the observation counts of the dataset. This layer ensures that model performance is evaluated under consistent conditions by maintaining a consistent data pattern and structure.

\textit{Method Layer} integrates dozens of different types of irregular time series forecasting methods, as well as multiple multivariate time series forecasting methods and time series foundation models. This layer features a unified model interface and scalability, ensuring consistency in evaluation and compatibility with different models. Subsequent work can build upon this foundation by writing simple and general-purpose interfaces to easily integrate their new models, thus enabling fair comparisons. This layer ensures the versatility and flexibility of BITS by providing support for a variety of models.

\textit{Evaluation Layer} includes various evaluation metrics, including commonly used error-based metrics and complementary non-error-based metrics, to enable more comprehensive evaluation. We also address the unreasonable lookback and prediction lengths in existing studies by setting customized and uniform lookback and prediction lengths for fair and effective evaluation. This layer comprehensively and effectively evaluates the performance of various models through multidimensional assessment across multiple metrics and more reasonable history and prediction lengths.


\textit{Reporting Layer} records experimental configurations, training and evaluation logs, runtime and resource usage, and environment information to ensure reproducibility and traceability. It also provides visualization tools for intuitive analysis of forecasting results. Moreover, BITS summarizes benchmark results across datasets and metrics and generates model rankings and leaderboards, enabling users to conveniently compare different methods and identify competitive models under diverse irregular time series forecasting scenarios.

Users only need to deploy their method at the method layer and configure the configuration file. Then BITS can automatically run the pipeline in Figure \ref{fig:ITFB}, enabling users to better understand, compare, and select irregular time series forecasting methods for specific irregular scenarios.

\vspace{-0.3cm}
\section{Experiments}
\label{sec: Experiments}
\vspace{-0.2cm}
\subsection{Experimental Setup}
\label{sec: experimental setup}

All experiments are conducted using PyTorch on an NVIDIA 3090 GPU. The models are trained with the MSE loss using the Adam optimizer. To ensure a consistent training objective across trainable baselines, we use MSE as the default loss function, while all other metrics are used only for evaluation and are not directly optimized during training, allowing us to assess model behavior under a unified optimization setting. For dataset splitting, we set it to 6:2:2 for training, validation, and test. The training, validation, and test sets are kept consistent for all models. We set customized lookback and prediction lengths for different datasets. The lookback horizons are set to 24 for PhysioNet and USHCN, 6000 for Human Activity and Pamap2, 336 for EPA-Air, 168 for ClusterTrace, 672 for CESNET, 400 for APTC, 730 for FNSPID, 1440 for Seabirds, 365 for GDELT. The prediction lengths are set to \{1, 6, 12\} for PhysioNet and USHCN, \{200, 1200, 3000\} for Human Activity, \{200, 1200, 3000\} for Pamap2, \{24, 72, 168\} for EPA-Air, \{24, 48, 72\} for ClusterTrace, \{24, 168, 336\} for CESNET, \{50, 100, 200\} for APTC, \{30, 180, 365\} for FNSPID, \{60, 360, 720\} for Seabirds, \{30, 90, 180\} for GDELT. For each method, we perform hyperparameter searches across multiple sets and report the best prediction performance. For splitting strategy, we adopt sample-aware splitting for PhysioNet, USHCN, APTC, and time-aware splitting for the remaining datasets.

\vspace{-0.2cm}
\subsection{Experimental Results}

\vspace{-0.1cm}
\subsubsection{Performance Comparison on Irregular Time Series Forecasting}

\begin{figure}[h]
    \centering
    \begin{subfigure}[t]{0.47\linewidth}
        \centering
        \includegraphics[width=\linewidth]{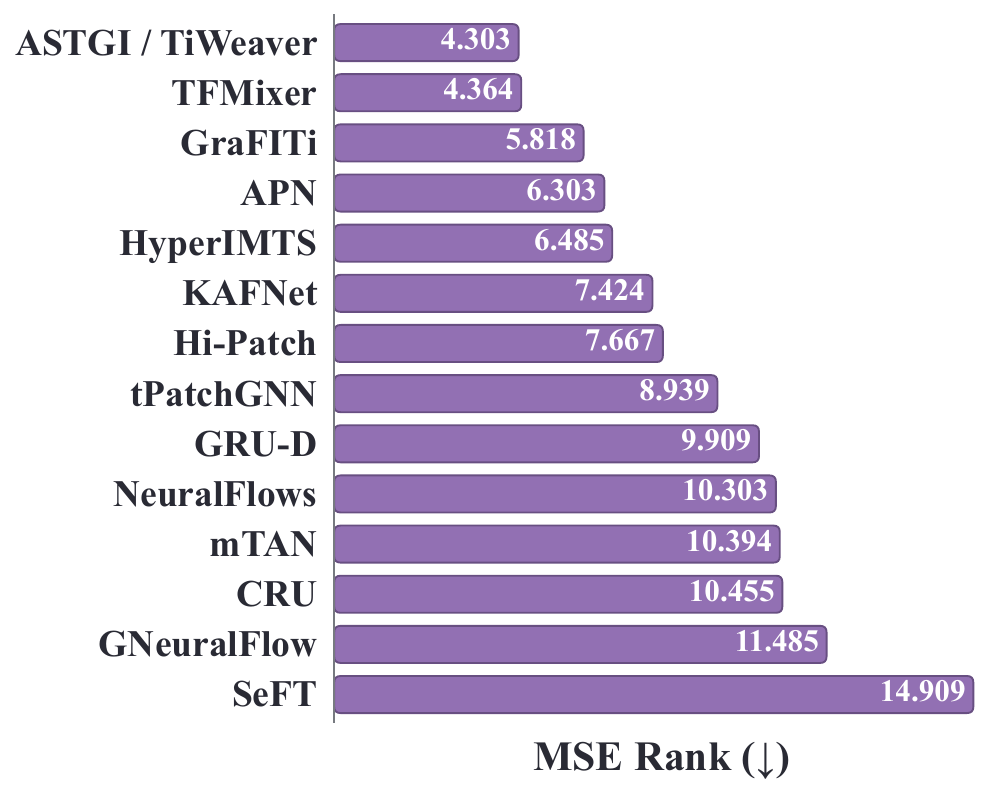}
        \label{aggregated mse rank}
    \end{subfigure}
    \hfill
    \begin{subfigure}[t]{0.47\linewidth}
        \centering
        \includegraphics[width=\linewidth]{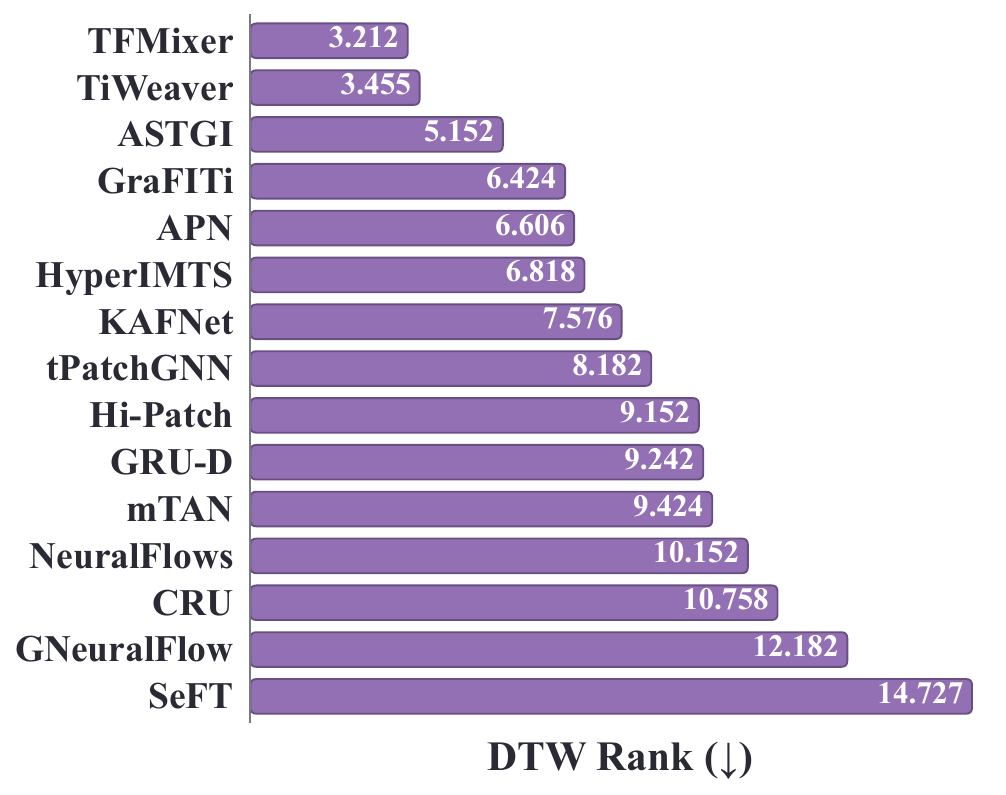}
        \label{aggregated dtw rank}
    \end{subfigure}

    \vspace{-0.5cm}
    \caption{Overall model rankings on BITS based on average per-task MSE and DTW ranks.}
    \label{fig:rank}
    \vspace{-0.1cm}
\end{figure}

We evaluate a wide range of models on 11 irregular time series datasets, reporting the average MSE over three prediction lengths in Table~\ref{tab:avg_mse_results} and Table~\ref{tab:avg_regular_tsfm_mse_results}, and ranking irregular models using two criteria in Figure~\ref{fig:rank}. Overall, irregular forecasting methods generally outperform regular forecasting models, demonstrating the benefits of explicitly handling irregular observations. Among them, TiWeaver, TFMixer and ASTGI demonstrate the strongest overall performance. TiWeaver and ASTGI achieve the lowest average MSE rank, and TFMixer ranks first in DTW. These results highlight the strong overall performance of temporal segmentation and aggregation methods and data-format transformation methods, suggesting that reorganizing irregular observations into informative temporal segments (e.g., patches) or structured representations (e.g., graphs and hypergraphs) is a promising direction for irregular forecasting. 

However, performance varies considerably within these categories. APN and tPatchGNN excel on the highly sparse PhysioNet, but tPatchGNN performs poorly on USHCN and EPA-Air, where TFMixer remains competitive, suggesting that temporal segmentation (e.g., patching) alone may be insufficient and that how information is integrated across temporal representations also matters. A similar disparity is observed among data-format transformation methods: ASTGI and GraFITi perform well overall, whereas SeFT falls behind. These observations indicate that transforming irregular observations into structured representations is beneficial, but the effectiveness of the resulting representation and subsequent dependency modeling remains crucial. Furthermore, the differences between the MSE and DTW rankings highlight the importance of incorporating non-error-based metrics for comprehensive evaluation.

Moreover, regular models occasionally outperform specialized irregular methods. DUET and SRSNet achieve the lowest average MSE on USHCN and EPA-Air, respectively, despite their high missing rates. This challenges the assumption that explicit irregularity modeling necessarily leads to better forecasting accuracy, which suggests that capturing predictive patterns remains essential, and that current irregular forecasting methods may not always translate their specialized designs into effective representations. Together, these findings highlight the importance of developing models that combine effective irregularity handling with robust temporal and cross-variable representation learning, rather than relying only on increasingly sophisticated irregularity-specific architectures. We further analyze the trade-off between forecasting performance and computational efficiency in terms of inference time and memory usage, as detailed in Appendix~\ref{sec: Model Efficiency Analysis}.

\begin{table}[t]
\centering
\caption{Average performance comparison of irregular time series forecasting models in terms of MSE ($\times 10^{-2}$) over three prediction lengths. Full results are provided in Appendix~\ref{sec: full results}.}
\label{tab:avg_mse_results}

\setlength{\tabcolsep}{2.2pt}
\renewcommand{\arraystretch}{1.10}

\makebox[\textwidth][c]{%
\resizebox{\textwidth}{!}{%
\begin{tabular}{c|cccccccccccccccc}
\toprule
\textbf{Dataset} &
\textbf{TFM} &
\textbf{TiW} &
\textbf{AST} &
\textbf{APN} &
\textbf{KAF} &
\textbf{HiP} &
\textbf{HIMTS} &
\textbf{GNF} &
\textbf{tPGNN} &
\textbf{GFT} &
\textbf{Warp} &
\textbf{CRU} &
\textbf{NF} &
\textbf{mTAN} &
\textbf{SeFT} &
\textbf{GRU-D} \\
\midrule

PhysioNet
& 0.558
& 0.567
& 0.518
& \underline{0.471}
& 0.567
& 0.551
& 0.588
& 1.040
& \textbf{0.468}
& 0.567
& 1.125
& 1.167
& 0.996
& 0.890
& 1.177
& 0.957 \\

Human Activity
& \textbf{0.363}
& 0.402
& 0.390
& \underline{0.366}
& 0.383
& 0.395
& 0.391
& 1.841
& 0.383
& 0.378
& 1.393
& 0.897
& 1.004
& 0.587
& 1.402
& 0.561 \\

USHCN
& \textbf{50.48}
& 56.08
& \underline{52.36}
& 54.86
& 64.47
& 58.32
& 68.96
& 59.13
& 69.21
& 59.58
& 53.19
& 63.49
& 57.26
& 68.15
& 76.41
& 68.96 \\

CESNET
& 2.327
& \textbf{2.156}
& \underline{2.160}
& 2.319
& 2.351
& 2.435
& 2.657
& 2.339
& 2.334
& 2.281
& 3.049
& 2.338
& 2.331
& 2.334
& 4.396
& 2.424 \\

Pamap2
& \textbf{0.599}
& 0.628
& 0.646
& 0.964
& 1.005
& 0.861
& \underline{0.621}
& 1.804
& 0.852
& 0.688
& -
& 1.297
& 1.550
& 0.872
& 1.874
& 1.775 \\

EPA-Air
& \underline{0.791}
& 0.813
& \textbf{0.778}
& 1.731
& 1.539
& 1.059
& 1.009
& 2.236
& 1.579
& 0.838
& 2.268
& 1.505
& 1.733
& 1.995
& 2.694
& 2.050 \\

ClusterTrace
& \underline{4.199}
& 4.332
& \textbf{4.177}
& 4.317
& 4.492
& 4.301
& 4.406
& 4.576
& 4.446
& 4.498
& 4.627
& 4.937
& 4.517
& 4.766
& 4.910
& 4.504 \\

APTC
& \textbf{0.381}
& \underline{0.381}
& 0.383
& 0.382
& 0.381
& 0.382
& 0.381
& 0.383
& 0.602
& 0.381
& 0.535
& 0.381
& 0.384
& 0.382
& 0.426
& 0.381 \\

FNSPID
& \underline{0.332}
& \textbf{0.300}
& 1.069
& 0.785
& 0.373
& 0.662
& 0.641
& 0.938
& 0.893
& 0.885
& 1.801
& 1.132
& 0.727
& 0.660
& 1.722
& 1.175 \\

GDELT
& 1.859
& \underline{1.829}
& \textbf{1.826}
& 1.841
& 1.837
& 1.955
& 1.849
& 1.865
& 1.859
& 1.953
& 2.307
& 1.864
& 1.843
& 1.918
& 2.720
& 1.837 \\

Seabirds
& 0.032
& \underline{0.026}
& \textbf{0.022}
& 0.081
& 0.097
& 0.098
& 0.049
& 0.134
& 0.117
& 0.075
& 0.117
& 0.114
& 0.170
& 0.116
& 0.141
& 0.164 \\

\bottomrule
\end{tabular}%
}%
}

\vspace{1pt}
\parbox{\textwidth}{%
\scriptsize
\raggedright
\textit{Note.} ``-'' indicates out-of-memory (OOM) even with the batch size set to 1.
\textit{Abbreviations.}
TFM: TFMixer;
TiW: TiWeaver;
AST: ASTGI;
KAF: KAFNet;
HiP: Hi-Patch;
HIMTS: HyperIMTS;
GNF: GNeuralFlow;
tPGNN: tPatchGNN;
GFT: GraFITi;
Warp: Warpformer;
NF: NeuralFlows.
}
\vspace{-0.7cm}
\end{table}

\vspace{-0.2cm}
\subsubsection{Irregular Time Series Forecasting with Time Series Foundation Model}

We further evaluate the generalization of time series foundation models (TSFMs) to irregular time series forecasting. As shown in Table~\ref{tab:avg_regular_tsfm_mse_results}, we compare three representative TSFMs under the zero-shot setting, reporting the average MSE over three prediction lengths. Detailed results are provided in Appendix~\ref{sec: full results}. Overall, TSFMs generally underperform specialized irregular forecasting methods, particularly on highly sparse datasets such as PhysioNet and USHCN. This suggests that large-scale pretraining alone may not be sufficient to effectively handle missing values, irregular sampling, and asynchronous variables. Among the evaluated TSFMs, MIRA remains competitive on CESNET, ClusterTrace, FNSPID, and Seabirds. Notably, ClusterTrace exhibits the highest sampling irregularity but a relatively low missing rate, whereas PhysioNet and USHCN suffer from severe missingness and complex missing patterns. This contrast suggests that irregular sampling and missing-data structures pose distinct challenges for zero-shot transfer, with the latter potentially creating a greater obstacle. These findings motivate irregularity-aware pretraining and adaptation strategies that explicitly account for missing observations and asynchronous variables.

\begin{table}[t]
\centering
\caption{Average performance comparison of time series forecasting models and TSFMs in terms of MSE ($\times 10^{-2}$) over three prediction lengths. Full results are provided in Appendix~\ref{sec: full results}.}
\label{tab:avg_regular_tsfm_mse_results}

\setlength{\tabcolsep}{2.2pt}
\renewcommand{\arraystretch}{1.10}

\makebox[\textwidth][c]{%
\resizebox{\textwidth}{!}{%
\begin{tabular}{c|cccccccccc}
\toprule
\textbf{Dataset} &
\textbf{Crossformer} &
\textbf{DLinear} &
\textbf{PatchTST} &
\textbf{TimesNet} &
\textbf{iTransformer} &
\textbf{DUET} &
\textbf{SRSNet} &
\textbf{MIRA} &
\textbf{Sundial} &
\textbf{Timer} \\
\midrule

PhysioNet
& \textbf{1.006}
& 5.819
& 3.844
& 4.537
& 3.847
& \underline{1.387}
& 1.413
& 8.740
& 17.292
& 12.601 \\

Human Activity
& 1.260
& 1.705
& 0.781
& 23.338
& 1.565
& \textbf{0.390}
& \underline{0.397}
& 9.425
& 29.647
& 23.550 \\

USHCN
& 73.40
& 76.55
& 76.22
& 75.55
& 76.03
& \textbf{49.01}
& \underline{55.05}
& 73.85
& 79.39
& 77.40 \\

CESNET
& 2.566
& 5.332
& 3.500
& 7.377
& 5.389
& \underline{2.276}
& 3.722
& \textbf{2.231}
& 5.132
& 5.133 \\

Pamap2
& 0.772
& 1.565
& 2.431
& 0.922
& 2.860
& \underline{0.559}
& \textbf{0.553}
& 5.711
& 11.063
& 11.620 \\

EPA-Air
& 1.955
& 2.808
& 1.082
& 1.293
& 1.163
& \underline{0.718}
& \textbf{0.672}
& 6.915
& 8.487
& 8.160 \\

ClusterTrace
& 6.355
& 8.142
& 7.100
& 9.309
& 7.697
& \underline{4.303}
& \textbf{4.269}
& 4.337
& 10.311
& 10.372 \\

APTC
& 0.380
& \textbf{0.375}
& 0.377
& \underline{0.375}
& 0.375
& 0.387
& 0.390
& 0.413
& 0.379
& 0.399 \\

FNSPID
& 3.213
& 4.813
& 0.881
& 6.872
& 2.811
& \underline{0.539}
& 5.626
& \textbf{0.439}
& 7.597
& 7.290 \\

GDELT
& \textbf{1.894}
& 15.587
& 4.760
& 9.921
& 7.743
& 1.969
& \underline{1.921}
& 3.884
& 16.282
& 15.502 \\

Seabirds
& 2.756
& 0.084
& 0.070
& 0.042
& 0.038
& \underline{0.027}
& \textbf{0.019}
& 0.038
& 0.045
& 0.116 \\

\bottomrule
\end{tabular}%
}%
}
\vspace{-0.3cm}
\end{table}

\vspace{-0.2cm}
\subsubsection{Case Study on the Limitation of Error-based Metrics}

\begin{figure}[h]
    \centering

    \begin{subfigure}[t]{0.47\linewidth}
        \centering
        \includegraphics[width=\linewidth]{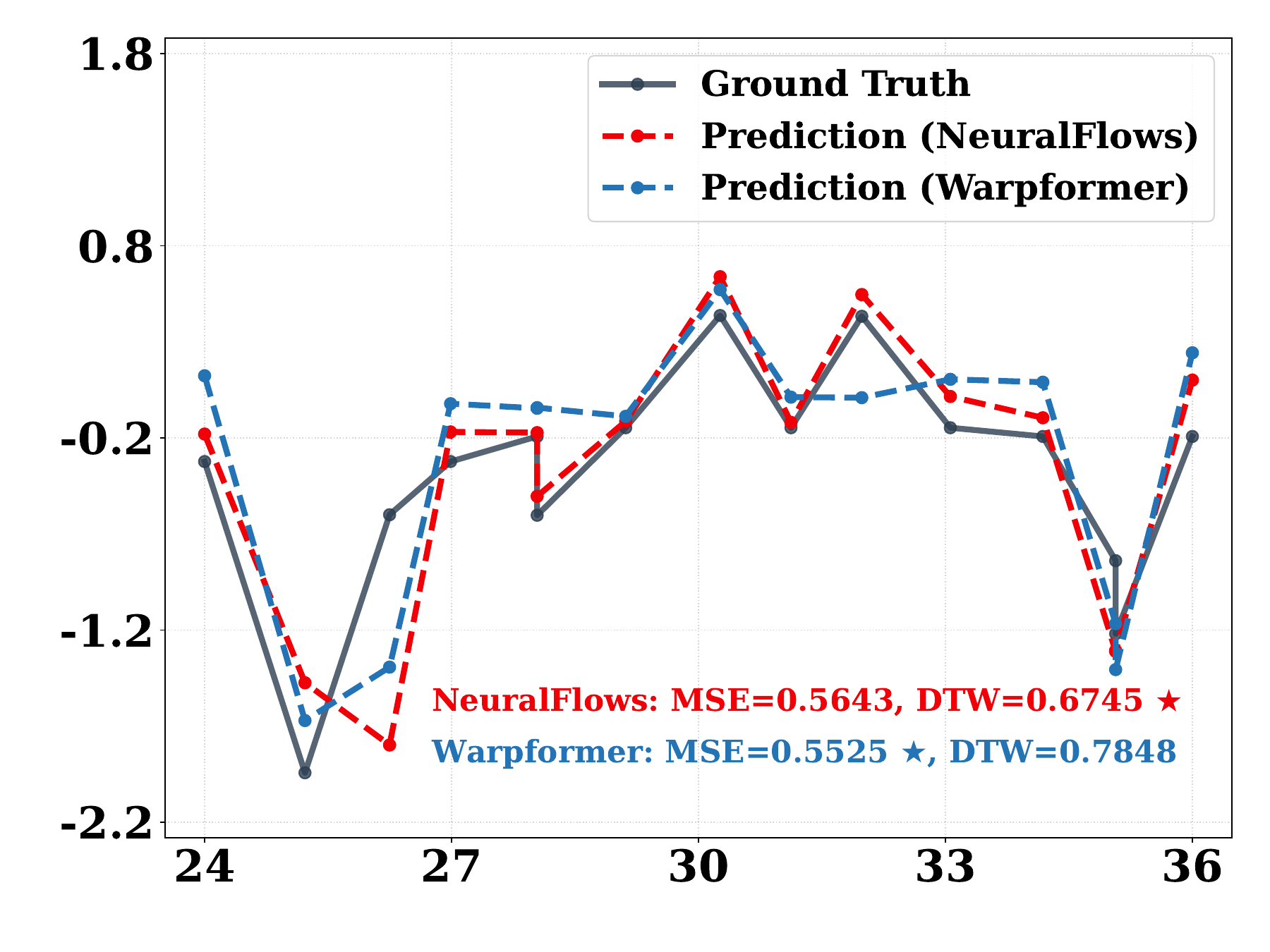}
        \vspace{-0.6cm}
        \caption{NeuralFlows vs. Warpformer}
        \label{NeuralFlows vs. Warpformer}
    \end{subfigure}
    \hfill
    \begin{subfigure}[t]{0.47\linewidth}
        \centering
        \includegraphics[width=\linewidth]{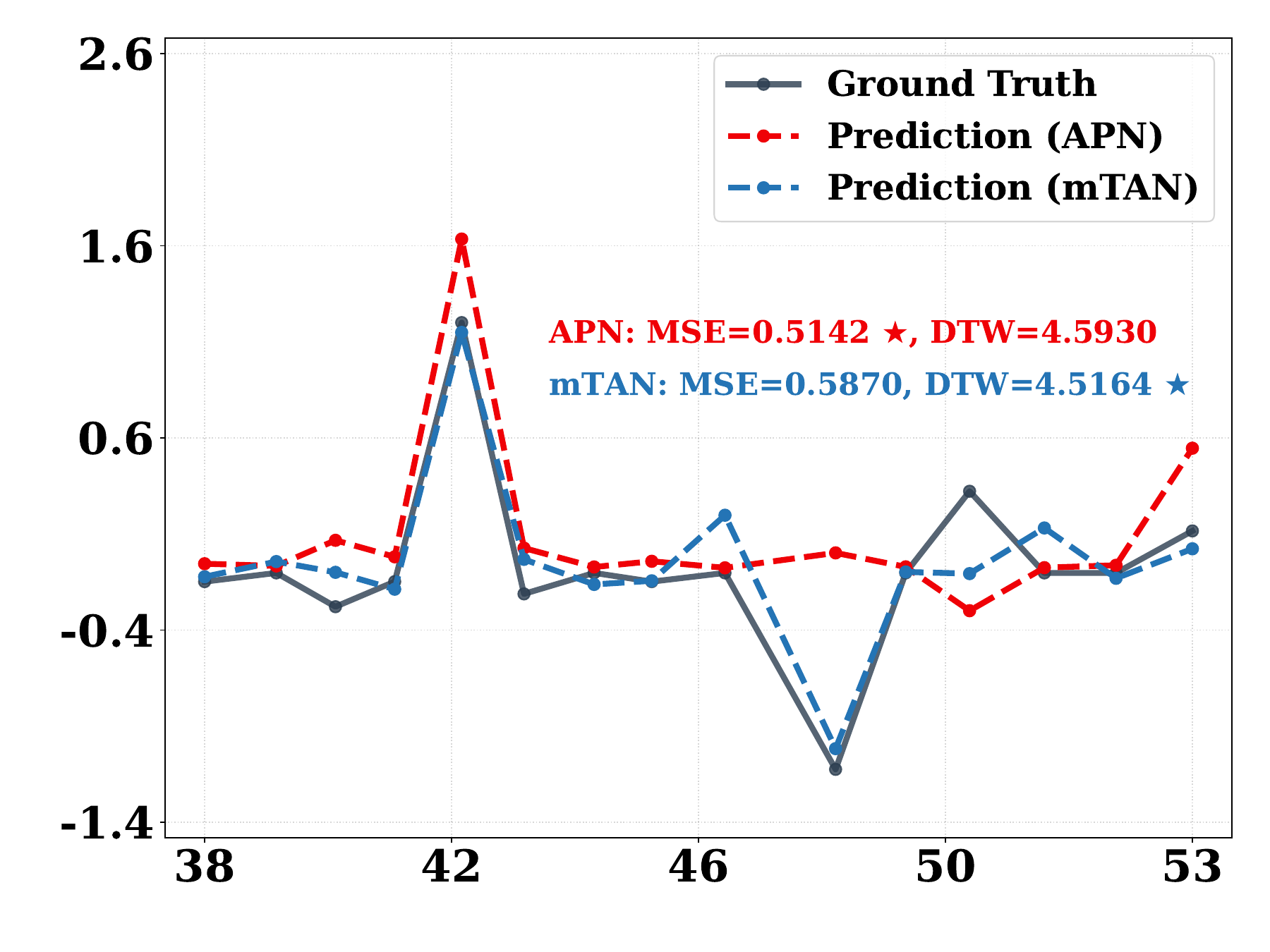}
        \vspace{-0.6cm}
        \caption{APN vs. mTAN}
        \label{APN vs. mTAN}
    \end{subfigure}

    \vspace{-0.3em}
    \caption{Prediction visualization on USHCN.}
    \label{fig:vis_pred}






\vspace{-0.5cm}
\end{figure}

To further illustrate the limitation of error-based metrics, we visualize two representative cases on USHCN in Figure~\ref{fig:vis_pred}. In Figure~\ref{NeuralFlows vs. Warpformer}, Warpformer achieves a lower MSE than NeuralFlows, while NeuralFlows obtains a lower DTW and better follows the local rising and falling patterns of the ground truth. A similar inconsistency is observed in Figure~\ref{APN vs. mTAN}: APN achieves a lower MSE than mTAN, while mTAN has a lower DTW and better captures several sharp temporal variations. These examples show that lower point-wise errors do not necessarily imply better temporal shape modeling. Therefore, non-error-based metrics provide complementary information for evaluating whether predictions preserve the underlying temporal dynamics.

\subsubsection{Comparison of six categories of irregular forecasting models}

\begin{figure}[h]
    \centering

    \begin{minipage}[t]{0.48\linewidth}
        \centering
        \begin{minipage}[c][5.0cm][c]{\linewidth}
            \centering
            \includegraphics[width=\linewidth]{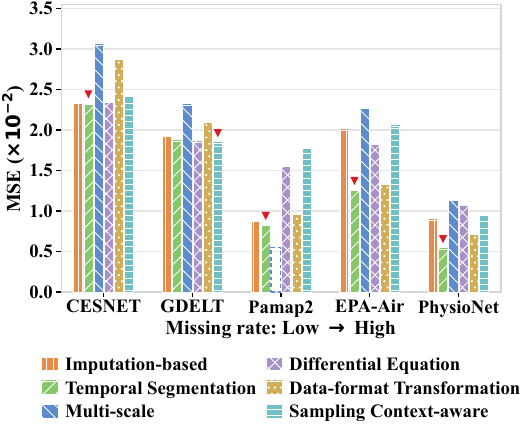}
        \end{minipage}
        \captionof{figure}{Comparison of six model categories across datasets with varying missing rates. Red triangles indicate the best-performing category. The missing bar on Pamap2 indicates OOM.}
        \label{fig:missing_rate_impact}
    \end{minipage}
    \hfill
    \begin{minipage}[t]{0.48\linewidth}
        \centering
        \begin{minipage}[c][5.0cm][c]{\linewidth}
            \centering
            \includegraphics[width=\linewidth]{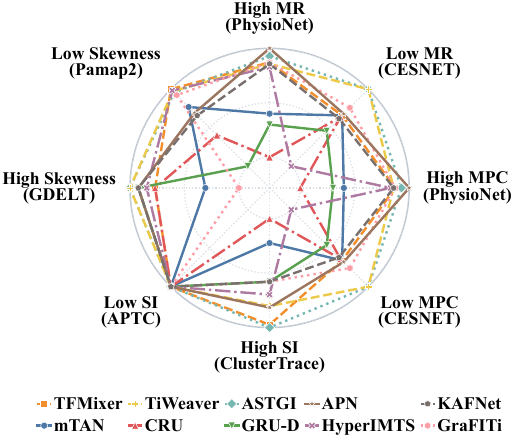}
        \end{minipage}
        \captionof{figure}{The MSE results for different irregularity characteristics. MR, MPC, and SI denote missing rate, missing pattern complexity, and sampling irregularity, respectively.}
        \label{fig:radarchart}
    \end{minipage}

\end{figure}

We compare six categories of irregular forecasting methods across different missing rates, as shown in Figure~\ref{fig:missing_rate_impact}. Temporal segmentation and aggregation methods achieve the best performance under most missing rates, and data-format transformation methods also perform competitively under high missing rates. Their advantages may stem from their ability to effectively utilize limited observations. For example, temporal aggregation consolidates sparse observations into informative local representations, while graph- and hypergraph-based transformations exploit complementary information across variables. Imputation-based and differential equation-based methods exhibit higher errors across different missing rates, potentially due to the difficulty of reconstructing missing values or estimating continuous dynamics from limited observations. These findings suggest that effectively utilizing available observations may be more beneficial than attempting to reconstruct fine-grained temporal information under severe missingness. In addition, the sampling context-aware modeling method achieves the best performance on GDELT. Sampling context-aware modeling (e.g., GRU-D) explicitly incorporates observation masks and time gaps, potentially allowing it to capture useful temporal information without complex representation transformations when observations are relatively abundant. This suggests that explicit sampling-context modeling can provide an effective alternative to temporal aggregation methods under certain irregularity conditions.

\vspace{-0.3cm}
\subsubsection{Performance on different irregularity characteristics}

We further evaluate ten representative models across different irregularity characteristics, as shown in Figure~\ref{fig:radarchart}. No model consistently performs best across all irregularity settings. APN achieves the best performance on PhysioNet with high missing rate and missing pattern complexity, whereas its advantage does not extend to datasets with other characteristics. This may benefit from its adaptive patching mechanism, which flexibly aggregates informative observations under severe sparsity. ASTGI performs strongly on CESNET, ClusterTrace, and GDELT, spanning low missingness, high sampling irregularity and high skewness. Its adaptive spatio-temporal graph construction may provide greater flexibility in capturing dependencies among irregular observations under these heterogeneous conditions. TiWeaver achieves the lowest MSE on CESNET and remains competitive on GDELT, suggesting the adaptability of its contextual patching strategy. Meanwhile, TFMixer achieves the best performance on Pamap2 with low skewness, despite ranking strongly but not uniformly best in the other settings. These complementary strengths indicate that different modeling strategies respond differently to irregularity characteristics, and strong overall performance does not imply universal superiority. These results highlight the importance of evaluating models across diverse irregular characteristics rather than relying on a limited number of datasets and irregularity settings.

\vspace{-0.3cm}
\section{Conclusion}
\label{sec: Conclusions}
\vspace{-0.2cm}

In this paper, we present BITS, a fair and comprehensive benchmark for irregular time series forecasting. BITS provides eleven multi-domain datasets with diverse irregularity characteristics, together with a standardized and scalable pipeline covering preprocessing, splitting, modeling, evaluation, and reporting. It further characterizes datasets from multiple irregularity dimensions and incorporates both error-based and non-error-based metrics for more comprehensive evaluation. Extensive experiments reveal that TiWeaver and TFMixer achieve consistently strong overall performance across multiple evaluation criteria, with ASTGI also demonstrating competitive results. Nevertheless, model performance varies across irregularity characteristics, and regular forecasting models can outperform specialized irregular methods in certain settings. TSFMs also show limitations under severe missingness and complex missing patterns, motivating irregularity-aware pretraining and adaptation. Our analysis also shows that error-based metrics alone may not fully reflect forecasting performance. Overall, BITS provides a reproducible and extensible evaluation platform for systematically comparing irregular time series forecasting methods.

\subsection*{AI use statement}



In this work, we did not use generative AI tools for tasks requiring mandatory disclosure, and the remaining required-disclosure tasks are not applicable to this work. Additionally, we used generative AI tools to assist with language polishing, improving the readability and organization of the manuscript, and identifying potentially relevant literature. We have reviewed all AI-assisted work and manually verified the resulting revisions and literature suggestions. We take responsibility for the final content of this work, including text, claims, or artifacts produced with the aid of generative AI.




\subsection*{Reproducibility statement}



We provide an anonymous implementation of BITS to facilitate reproducibility. The benchmark includes standardized data preprocessing, model integration, evaluation, and reporting procedures. Detailed dataset descriptions and preprocessing settings are provided in Appendix~\ref{datasets_appendix}, evaluation metrics are described in Appendix~\ref{appendix_metrics}, irregularity characteristics are defined in Appendix~\ref{data_characteristics_appendix}, and complete experimental results are reported in Appendix~\ref{sec: full results}. Experimental settings, including dataset splitting strategies, lookback lengths, prediction lengths, and training configurations, are described in Sections~\ref{sec: ITFB Benchmark Framework} and~\ref{sec: experimental setup}. The source code is available at the anonymous repository provided in the paper.



\bibliography{iclr2027_conference}

\begin{thebibliography}{44}
\providecommand{\natexlab}[1]{#1}
\providecommand{\url}[1]{\texttt{#1}}
\expandafter\ifx\csname urlstyle\endcsname\relax
  \providecommand{\doi}[1]{doi: #1}\else
  \providecommand{\doi}{doi: \begingroup \urlstyle{rm}\Url}\fi

\bibitem[Bilos et~al.(2021)Bilos, Sommer, Rangapuram, Januschowski, and G{\"{u}}nnemann]{DBLP:conf/nips/BilosSRJG21}
Marin Bilos, Johanna Sommer, Syama~Sundar Rangapuram, Tim Januschowski, and Stephan G{\"{u}}nnemann.
\newblock Neural flows: Efficient alternative to neural odes.
\newblock In \emph{NeurIPS}, pp.\  21325--21337, 2021.

\bibitem[Browning et~al.(2018)Browning, Bolton, Owen, Shoji, Guilford, and Freeman]{browning2018predicting}
Ella Browning, Mark Bolton, Ellie Owen, Akiko Shoji, Tim Guilford, and Robin Freeman.
\newblock Predicting animal behaviour using deep learning: Gps data alone accurately predict diving in seabirds.
\newblock \emph{Methods in Ecology and Evolution}, 9\penalty0 (3):\penalty0 681--692, 2018.

\bibitem[Bui et~al.(2024)Bui, Warrier, and Gupta]{bui2024benchmarking}
Hung Bui, Harikrishna Warrier, and Yogesh Gupta.
\newblock Benchmarking with mimic-iv, an irregular, spare clinical time series dataset.
\newblock \emph{arXiv preprint arXiv:2401.15290}, 2024.

\bibitem[Chang et~al.(2025)Chang, Hwang, Shi, Wang, Wang, Peng, and Chen]{DBLP:conf/nips/ChangHSWWPC25}
Ching Chang, Jeehyun Hwang, Yidan Shi, Haixin Wang, Wei Wang, Wen{-}Chih Peng, and Tien{-}Fu Chen.
\newblock Time-imm: {A} dataset and benchmark for irregular multimodal multivariate time series.
\newblock In \emph{{NeurIPS}}, 2025.

\bibitem[Che et~al.(2016)Che, Purushotham, Cho, Sontag, and Liu]{DBLP:journals/corr/ChePCSL16}
Zhengping Che, Sanjay Purushotham, Kyunghyun Cho, David~A. Sontag, and Yan Liu.
\newblock Recurrent neural networks for multivariate time series with missing values.
\newblock \emph{CoRR}, abs/1606.01865, 2016.

\bibitem[de~Souza(2018)]{DBLP:journals/eaai/Souza18}
Vin{\'{\i}}cius M.~A. de~Souza.
\newblock Asphalt pavement classification using smartphone accelerometer and complexity invariant distance.
\newblock \emph{Eng. Appl. Artif. Intell.}, 74:\penalty0 198--211, 2018.

\bibitem[Dong et~al.(2024)Dong, Fan, and Peng]{DBLP:conf/kdd/DongFP24}
Zihan Dong, Xinyu Fan, and Zhiyuan Peng.
\newblock {FNSPID:} {A} comprehensive financial news dataset in time series.
\newblock In Ricardo Baeza{-}Yates and Francesco Bonchi (eds.), \emph{Proceedings of the 30th {ACM} {SIGKDD} Conference on Knowledge Discovery and Data Mining, {KDD} 2024, Barcelona, Spain, August 25-29, 2024}, pp.\  4918--4927. {ACM}, 2024.
\newblock \doi{10.1145/3637528.3671629}.
\newblock URL \url{https://doi.org/10.1145/3637528.3671629}.

\bibitem[Godahewa et~al.(2021)Godahewa, Bergmeir, Webb, Hyndman, and Montero{-}Manso]{DBLP:conf/nips/GodahewaBWHM21}
Rakshitha Godahewa, Christoph Bergmeir, Geoffrey~I. Webb, Rob~J. Hyndman, and Pablo Montero{-}Manso.
\newblock Monash time series forecasting archive.
\newblock In \emph{Proceedings of the Neural Information Processing Systems Track on Datasets and Benchmarks 1, NeurIPS Datasets and Benchmarks 2021, December 2021, virtual}, 2021.

\bibitem[Harutyunyan et~al.(2019)Harutyunyan, Khachatrian, Kale, Ver~Steeg, and Galstyan]{Harutyunyan2019}
Hrayr Harutyunyan, Hrant Khachatrian, David~C. Kale, Greg Ver~Steeg, and Aram Galstyan.
\newblock Multitask learning and benchmarking with clinical time series data.
\newblock \emph{Scientific Data}, 6\penalty0 (1):\penalty0 96, 2019.

\bibitem[Horn et~al.(2020)Horn, Moor, Bock, Rieck, and Borgwardt]{DBLP:conf/icml/HornMBRB20}
Max Horn, Michael Moor, Christian Bock, Bastian Rieck, and Karsten~M. Borgwardt.
\newblock Set functions for time series.
\newblock In \emph{{ICML}}, volume 119, pp.\  4353--4363, 2020.

\bibitem[Koumar et~al.(2025)Koumar, Hynek, {\v{C}}ejka, and {\v{S}}i{\v{s}}ka]{koumar2025cesnet}
Josef Koumar, Karel Hynek, Tom{\'a}{\v{s}} {\v{C}}ejka, and Pavel {\v{S}}i{\v{s}}ka.
\newblock Cesnet-timeseries24: Time series dataset for network traffic anomaly detection and forecasting.
\newblock \emph{Scientific Data}, 12\penalty0 (1):\penalty0 338, 2025.

\bibitem[Leetaru \& Schrodt(2013)Leetaru and Schrodt]{leetaru2013gdelt}
Kalev Leetaru and Philip~A Schrodt.
\newblock Gdelt: Global data on events, location, and tone, 1979--2012.
\newblock In \emph{ISA annual convention}, volume~2, pp.\  1--49, 2013.

\bibitem[Li et~al.(2025{\natexlab{a}})Li, Luo, Liu, Zheng, Lv, and Ma]{DBLP:conf/icml/LiL0ZL025}
Boyuan Li, Yicheng Luo, Zhen Liu, Junhao Zheng, Jianming Lv, and Qianli Ma.
\newblock Hyperimts: Hypergraph neural network for irregular multivariate time series forecasting.
\newblock In \emph{{ICML}}, 2025{\natexlab{a}}.

\bibitem[Li et~al.(2025{\natexlab{b}})Li, Deng, Xu, Feng, Schlegel, Huang, Sun, Sun, Yang, Yu, and Bian]{DBLP:conf/nips/LiDXFSHSSYYB25}
Hao Li, Bowen Deng, Chang Xu, Zhiyuan Feng, Viktor Schlegel, Yu{-}Hao Huang, Yizheng Sun, Jingyuan Sun, Kailai Yang, Yiyao Yu, and Jiang Bian.
\newblock {MIRA:} medical time series foundation model for real-world health data.
\newblock In \emph{{NeurIPS}}, 2025{\natexlab{b}}.

\bibitem[Li et~al.(2025{\natexlab{c}})Li, Qiu, Chen, Wang, Cheng, Shu, Hu, Guo, Zhou, Jensen, et~al.]{li2025tsfm}
Zhe Li, Xiangfei Qiu, Peng Chen, Yihang Wang, Hanyin Cheng, Yang Shu, Jilin Hu, Chenjuan Guo, Aoying Zhou, Christian~S Jensen, et~al.
\newblock Tsfm-bench: A comprehensive and unified benchmark of foundation models for time series forecasting.
\newblock In \emph{{KDD}}, pp.\  5595--5606, 2025{\natexlab{c}}.

\bibitem[Li et~al.(2026)Li, Tian, Miao, Lei, Guo, and Yang]{DBLP:conf/kdd/LiTMLGY26}
Zhe Li, Jindong Tian, Hao Miao, Zhi Lei, Chenjuan Guo, and Bin Yang.
\newblock Tiweaver: Unified temporal dynamics modeling via contextual patching.
\newblock In \emph{{KDD}}, pp.\  2933--2944, 2026.

\bibitem[Liang et~al.(2022)Liang, Shao, Wang, Zhang, Sun, and Xu]{liang2022basicts}
Yubo Liang, Zezhi Shao, Fei Wang, Zhao Zhang, Tao Sun, and Yongjun Xu.
\newblock Basicts: An open source fair multivariate time series prediction benchmark.
\newblock In \emph{International symposium on benchmarking, measuring and optimization}, pp.\  87--101, 2022.

\bibitem[Liu et~al.(2025{\natexlab{a}})Liu, Qiu, Wu, Li, Guo, Hu, and Yang]{DBLP:journals/corr/abs-2505-11250}
Xvyuan Liu, Xiangfei Qiu, Xingjian Wu, Zhengyu Li, Chenjuan Guo, Jilin Hu, and Bin Yang.
\newblock Rethinking irregular time series forecasting: {A} simple yet effective baseline.
\newblock \emph{CoRR}, abs/2505.11250, 2025{\natexlab{a}}.

\bibitem[Liu et~al.(2026)Liu, Qiu, Cheng, Wu, Guo, Yang, and Hu]{liu2026astgi}
Xvyuan Liu, Xiangfei Qiu, Hanyin Cheng, Xingjian Wu, Guo Guo, Bin Yang, and Jilin Hu.
\newblock Astgi: Adaptive spatio-temporal graph interactions for irregular multivariate time series forecasting.
\newblock In \emph{{ICLR}}, volume 2026, pp.\  23940--23958, 2026.

\bibitem[Liu et~al.(2024{\natexlab{a}})Liu, Hu, Zhang, Wu, Wang, Ma, and Long]{DBLP:conf/iclr/LiuHZWWML24}
Yong Liu, Tengge Hu, Haoran Zhang, Haixu Wu, Shiyu Wang, Lintao Ma, and Mingsheng Long.
\newblock itransformer: Inverted transformers are effective for time series forecasting.
\newblock In \emph{{ICLR}}, 2024{\natexlab{a}}.

\bibitem[Liu et~al.(2024{\natexlab{b}})Liu, Zhang, Li, Huang, Wang, and Long]{liu2024timer}
Yong Liu, Haoran Zhang, Chenyu Li, Xiangdong Huang, Jianmin Wang, and Mingsheng Long.
\newblock Timer: generative pre-trained transformers are large time series models.
\newblock In \emph{{ICML}}, pp.\  32369--32399, 2024{\natexlab{b}}.

\bibitem[Liu et~al.(2025{\natexlab{b}})Liu, Qin, Shi, Chen, Yang, Huang, Wang, and Long]{DBLP:conf/icml/LiuQSCY00L25}
Yong Liu, Guo Qin, Zhiyuan Shi, Zhi Chen, Caiyin Yang, Xiangdong Huang, Jianmin Wang, and Mingsheng Long.
\newblock Sundial: {A} family of highly capable time series foundation models.
\newblock In \emph{{ICML}}, 2025{\natexlab{b}}.

\bibitem[Luo et~al.(2025)Luo, Zhang, Liu, and Ma]{DBLP:conf/icml/LuoZ0025}
Yicheng Luo, Bowen Zhang, Zhen Liu, and Qianli Ma.
\newblock Hi-patch: Hierarchical patch {GNN} for irregular multivariate time series.
\newblock In \emph{{ICML}}, 2025.

\bibitem[Menne et~al.(2015)Menne, Williams~Jr, and Vose]{menne2015long}
Matthew~J Menne, Claude~N Williams~Jr, and Russell~S Vose.
\newblock Long-term daily and monthly climate records from stations across the contiguous united states (us historical climatology network).
\newblock Technical report, Environmental System Science Data Infrastructure for a Virtual Ecosystem; CDIAC, 2015.

\bibitem[Mercatali et~al.(2024)Mercatali, Freitas, and Chen]{DBLP:conf/nips/MercataliFC24}
Giangiacomo Mercatali, Andr{\'{e}} Freitas, and Jie Chen.
\newblock Graph neural flows for unveiling systemic interactions among irregularly sampled time series.
\newblock In \emph{NeurIPS}, 2024.

\bibitem[Nie et~al.(2023)Nie, Nguyen, Sinthong, and Kalagnanam]{DBLP:conf/iclr/NieNSK23}
Yuqi Nie, Nam~H. Nguyen, Phanwadee Sinthong, and Jayant Kalagnanam.
\newblock A time series is worth 64 words: Long-term forecasting with transformers.
\newblock In \emph{{ICLR}}, 2023.

\bibitem[Qiu et~al.(2024)Qiu, Hu, Zhou, Wu, Du, Zhang, Guo, Zhou, Jensen, Sheng, et~al.]{qiu2024tfb}
Xiangfei Qiu, Jilin Hu, Lekui Zhou, Xingjian Wu, Junyang Du, Buang Zhang, Chenjuan Guo, Aoying Zhou, Christian~S Jensen, Zhenli Sheng, et~al.
\newblock Tfb: Towards comprehensive and fair benchmarking of time series forecasting methods.
\newblock \emph{Proceedings of the VLDB Endowment}, 17\penalty0 (9):\penalty0 2363--2377, 2024.

\bibitem[Qiu et~al.(2025)Qiu, Wu, Lin, Guo, Hu, and Yang]{DBLP:conf/kdd/QiuW0GH025}
Xiangfei Qiu, Xingjian Wu, Yan Lin, Chenjuan Guo, Jilin Hu, and Bin Yang.
\newblock {DUET:} dual clustering enhanced multivariate time series forecasting.
\newblock In \emph{{KDD}}, pp.\  1185--1196, 2025.

\bibitem[Qiu et~al.(2026)Qiu, Yan, Liu, Wu, and Hu]{qiu2026bridging}
Xiangfei Qiu, Kangjia Yan, Xvyuan Liu, Xingjian Wu, and Jilin Hu.
\newblock Bridging time and frequency: A joint modeling framework for irregular multivariate time series forecasting.
\newblock In \emph{ICML}, 2026.

\bibitem[Reiss \& Stricker(2012)Reiss and Stricker]{DBLP:conf/iswc/ReissS12}
Attila Reiss and Didier Stricker.
\newblock Introducing a new benchmarked dataset for activity monitoring.
\newblock In \emph{{ISWC}}, pp.\  108--109, 2012.

\bibitem[Schirmer et~al.(2022)Schirmer, Eltayeb, Lessmann, and Rudolph]{DBLP:conf/icml/SchirmerELR22}
Mona Schirmer, Mazin Eltayeb, Stefan Lessmann, and Maja Rudolph.
\newblock Modeling irregular time series with continuous recurrent units.
\newblock In \emph{{ICML}}, volume 162, pp.\  19388--19405, 2022.

\bibitem[Shao et~al.(2025)Shao, Wang, Xu, Wei, Yu, Zhang, Yao, Sun, Jin, Cao, Cong, Jensen, and Cheng]{DBLP:journals/tkde/ShaoWXWYZYSJCCJC25}
Zezhi Shao, Fei Wang, Yongjun Xu, Wei Wei, Chengqing Yu, Zhao Zhang, Di~Yao, Tao Sun, Guangyin Jin, Xin Cao, Gao Cong, Christian~S. Jensen, and Xueqi Cheng.
\newblock Exploring progress in multivariate time series forecasting: Comprehensive benchmarking and heterogeneity analysis.
\newblock \emph{{IEEE} Trans. Knowl. Data Eng.}, 37\penalty0 (1):\penalty0 291--305, 2025.

\bibitem[Shukla \& Marlin(2021)Shukla and Marlin]{DBLP:conf/iclr/ShuklaM21}
Satya~Narayan Shukla and Benjamin~M. Marlin.
\newblock Multi-time attention networks for irregularly sampled time series.
\newblock In \emph{{ICLR}}, 2021.

\bibitem[Silva et~al.(2012)Silva, Moody, Scott, Celi, and Mark]{silva2012predicting}
Ikaro Silva, George Moody, Daniel~J Scott, Leo~A Celi, and Roger~G Mark.
\newblock Predicting in-hospital mortality of icu patients: The physionet computing in cardiology challenge 2012.
\newblock \emph{Computing in cardiology}, 39:\penalty0 245, 2012.

\bibitem[Wang et~al.(2026)Wang, Wu, Dong, Liu, Wang, Long, and Wang]{wang2026deep}
Yuxuan Wang, Haixu Wu, Jiaxiang Dong, Yong Liu, Chen Wang, Mingsheng Long, and Jianmin Wang.
\newblock Deep time series models: A comprehensive survey and benchmark.
\newblock \emph{IEEE Transactions on Pattern Analysis and Machine Intelligence}, 2026.

\bibitem[Weng et~al.(2022)Weng, Xiao, Yu, Wang, Wang, He, Li, Zhang, Lin, and Ding]{DBLP:conf/nsdi/WengXYWWHLZLD22}
Qizhen Weng, Wencong Xiao, Yinghao Yu, Wei Wang, Cheng Wang, Jian He, Yong Li, Liping Zhang, Wei Lin, and Yu~Ding.
\newblock Mlaas in the wild: Workload analysis and scheduling in large-scale heterogeneous {GPU} clusters.
\newblock In \emph{{NSDI}}, pp.\  945--960, 2022.

\bibitem[Wu et~al.(2023)Wu, Hu, Liu, Zhou, Wang, and Long]{wu2023timesnet}
Haixu Wu, Tengge Hu, Yong Liu, Hang Zhou, Jianmin Wang, and Mingsheng Long.
\newblock Timesnet: Temporal 2d-variation modeling for general time series analysis.
\newblock In \emph{{ICLR}}, 2023.

\bibitem[Wu et~al.(2025)Wu, Qiu, Cheng, Li, Hu, Guo, and Yang]{DBLP:conf/nips/WuQCLHGY25}
Xingjian Wu, Xiangfei Qiu, Hanyin Cheng, Zhengyu Li, Jilin Hu, Chenjuan Guo, and Bin Yang.
\newblock Enhancing time series forecasting through selective representation spaces: {A} patch perspective.
\newblock In \emph{{NeurIPS}}, 2025.

\bibitem[Yalavarthi et~al.(2024)Yalavarthi, Madhusudhanan, Scholz, Ahmed, Burchert, Jawed, Born, and Schmidt{-}Thieme]{DBLP:conf/aaai/YalavarthiMSABJ24}
Vijaya~Krishna Yalavarthi, Kiran Madhusudhanan, Randolf Scholz, Nourhan Ahmed, Johannes Burchert, Shayan Jawed, Stefan Born, and Lars Schmidt{-}Thieme.
\newblock Grafiti: Graphs for forecasting irregularly sampled time series.
\newblock In \emph{{AAAI}}, pp.\  16255--16263, 2024.

\bibitem[Zeng et~al.(2023)Zeng, Chen, Zhang, and Xu]{DBLP:conf/aaai/ZengCZ023}
Ailing Zeng, Muxi Chen, Lei Zhang, and Qiang Xu.
\newblock Are transformers effective for time series forecasting?
\newblock In \emph{{AAAI}}, pp.\  11121--11128, 2023.

\bibitem[Zhang et~al.(2023)Zhang, Zheng, Cao, Bian, and Li]{DBLP:conf/kdd/ZhangZCBL23}
Jiawen Zhang, Shun Zheng, Wei Cao, Jiang Bian, and Jia Li.
\newblock Warpformer: {A} multi-scale modeling approach for irregular clinical time series.
\newblock In \emph{{SIGKDD}}, pp.\  3273--3285, 2023.

\bibitem[Zhang et~al.(2024)Zhang, Yin, Liu, Zhou, and Xiong]{DBLP:conf/icml/ZhangYL0024}
Weijia Zhang, Chenlong Yin, Hao Liu, Xiaofang Zhou, and Hui Xiong.
\newblock Irregular multivariate time series forecasting: {A} transformable patching graph neural networks approach.
\newblock In \emph{{ICML}}, 2024.

\bibitem[Zhang \& Yan(2023)Zhang and Yan]{DBLP:conf/iclr/ZhangY23}
Yunhao Zhang and Junchi Yan.
\newblock Crossformer: Transformer utilizing cross-dimension dependency for multivariate time series forecasting.
\newblock In \emph{{ICLR}}, 2023.

\bibitem[Zhou et~al.(2026)Zhou, Huang, Wang, Wu, Kwok, and Liang]{DBLP:conf/aaai/ZhouHWWKL26}
Ziyu Zhou, Yiming Huang, Yanyun Wang, Yuankai Wu, James Kwok, and Yuxuan Liang.
\newblock Revitalizing canonical pre-alignment for irregular multivariate time series forecasting.
\newblock In \emph{{AAAI}}, pp.\  29115--29123, 2026.

\end{thebibliography}
\bibliographystyle{iclr2027_conference}

\appendix
\section{Appendix}
\subsection{Preliminaries}
\label{sec:preliminaries}

\subsubsection{Irregular Time Series}

Let a multivariate time series be defined as $\mathcal{X}=\{(t_i,\mathbf{x}_i)\}_{i=1}^N$, where $t_i \in \mathbb{R}^+$ denotes the timestamp of the $i$-th observation and $\mathbf{x}_i \in \mathbb{R}^D$ is a D-dimensional measurement vector. A time series is regular if timestamps are uniformly spaced, (i.e., $t_{i+1}- t_i = \Delta t, \forall i$). In contrast, a time series is irregular when the intervals are non-uniform: $t_{i+1}- t_i \ne \Delta t$, or $t_{i+1} - t_i \in \mathbb{R}^+$ varies significantly.

Irregularity also manifests in two common forms:

\textbf{Variable-level asynchronous sampling.} Each variable $x^{(d)}$ may have its own observation times:
\begin{equation}
    \mathcal{X}^{(d)}= \{(t_i^{(d)},x_i^{(d)})\}_{i=1}^{N_d}, t_i^{(d)} \ne t_j^{(d^\prime)}.
\end{equation}

\textbf{Instance-level missingness and sparsity.} For a shared timeline $t_1, ..., t_T$, some variables may be missing at certain time steps. A masking matrix $\mathbf{M} \in \{0,1\}^{T \times D}$ is typically defined as
\begin{equation}
    M_{t,d} = 
        \begin{cases}
        1, & \text{if } x_{t,d} \text{ is observed},\\[4pt]
        0, & \text{otherwise}.
        \end{cases}
\end{equation}
Irregular time series arise naturally in domains such as medicine, geoscience, and industrial monitoring due to asynchronous sensors, event-driven sampling, and device failures.

\subsubsection{Irregular Time Series Forecasting}
Given an irregular multivariate time series $\mathcal{X}_{1:T}= \{(t_i,\mathbf{x_i})\}_{i=1}^T$, the goal of irregular time series forecasting (ITSF) is to predict future values over a horizon $H$:
\begin{equation}
    \mathcal{X}_{T+1:T+H}= f_\theta(\mathcal{X}_{1:T}, \mathbf{M}_{1:T}, \Delta_{t_{1:T}}),
\end{equation}
where $f_\theta$ is the forecasting model, $\mathbf{M}_{1:T}$ contains observation masks, and $\Delta t_{1:T}$ captures time gaps between observations: $\Delta t_i= t_i- t_{i-1}$.

Different from regular forecasting, ITSF models must explicitly handle non-uniform temporal intervals. Two main challenges arise:

\textbf{Temporal Encoding under Irregularity.} Models need to account for non-uniform temporal intervals, for example, by feeding time gaps into the model to avoid misleading assumptions of uniform sampling: $\mathbf{z}_i= g(\mathbf{x}_i, \Delta t_i)$, where $g(\cdot)$ is a temporal encoding function (e.g., exponential decay, continuous-time embeddings, or neural ODE dynamics).

\textbf{Alignment Across Variables.} When variables are asynchronously sampled, the model must align their representations: $\mathbf{h}_t= A(\{(t_i^{(d)}, x_i^{(d)}): t_i^{(d)} \le t\})$, where $A(\cdot)$ aggregates information with respect to time.

\subsection{Datasets}
\label{datasets_appendix}
We collect a series of open-source irregular time series datasets that cover a wide range of domains and characteristics for irregular time series forecasting. Detailed descriptions of these datasets are provided below.

\textbf{PhysioNet}~\citep{silva2012predicting}. This dataset includes 12000 IMTS from different patients, each with 41 clinical signals collected irregularly during the first 48 hours of ICU admission. We use the first 24 hours as the observed data to predict the queried values in the subsequent 1/6/12 hours.


\textbf{Human Activity}. This dataset consists of 12 irregularly measured 3D positional variables from sensors worn on the ankles, belts, and chests of five individuals performing various activities. We chunk the original time series into 25 IMTS. The first 6000 milliseconds are used as observed data to predict the sensor positions for the next 200/1200/3000 milliseconds.

\textbf{USHCN}~\citep{menne2015long}. The USHCN dataset includes over 150 years of climate data from multiple U.S. stations, covering 5 climate variables. We focus on data from 1114 stations between 1996 and 2000, resulting in 1114 IMTS. Each instance uses data from the previous 24 months to predict the next 1/6/12 month’s climate conditions.

\textbf{CESNET}~\citep{koumar2025cesnet}. This dataset comprises network traffic data collected from the CESNET academic network, representing a large-scale collection of flow-level statistics. It contains 11 IMTS derived from millions of network flows, where each instance includes 5 features such as packet counts, byte counts, and inter-arrival times recorded over varying observation windows. Following standard protocol for network traffic analysis, the flows are processed into discrete intervals. We use the first 672 hours as the observed sequence to predict the traffic characteristics and flow types in the subsequent 24/168/336 hours.

\textbf{Pamap2}~\citep{DBLP:conf/iswc/ReissS12}. The Pamap2 dataset contains data from multiple participants performing different physical activities (such as walking, cycling, and Nordic walking) while wearing three inertial measurement units and a heart rate monitor. The raw sensory data includes multiple dimensions of 3D-acceleration, gyroscope, magnetometer, and temperature readings, along with heart rate. To adapt it for forecasting, the continuous recordings are segmented and 2 IMTS are used, each representing a specific activity window. We select 4 variables for irregular time series forecasting. The first 6000 milliseconds are utilized as observed data to predict the sensor readings for the following 200/1200/3000 milliseconds. 

\textbf{EPA-Air}. This dataset is derived from the Air Quality System (AQS), the U.S. Environmental Protection Agency’s repository of ambient air quality data. It consists of hourly and daily measurements from multiple monitoring stations across the United States, covering several criteria air pollutants (such as PM2.5, Ozone, and CO) and various meteorological variables. Due to sensor maintenance and varying sampling frequencies, the records form a large-scale collection of IMTS. To simulate real-world environmental monitoring and forecasting, the data is typically organized into thousands of time series instances based on station locations. We select 8 IMTS and 4 variables for irregular time series forecasting. We use the measurements from the previous 336 hours as the observed data to predict the pollutant concentrations and Air Quality Index (AQI) for the following 24/72/168 hours.

\textbf{ClusterTrace}~\citep{DBLP:conf/nsdi/WengXYWWHLZLD22}. The ClusterTrace dataset provides a detailed trace of a production cluster containing multiple machines over a period of time. It captures complex workloads and resource utilization metrics, including CPU, memory, and disk usage for various tasks and containers. Due to the dynamic scheduling and heterogeneous nature of cloud workloads, the resource metrics are recorded as IMTS across thousands of unique job instances. Following common practices in cluster management research, the data is processed to focus on temporal resource demands. We select 3 IMTS and 3 variables for irregular time series forecasting. We use the resource usage data from the first 168 hours as the observed sequence to predict the workload intensity and resource requirements for the subsequent 24/48/72 hours.

\textbf{APTC}~\citep{DBLP:journals/eaai/Souza18}. This dataset contains three-axis acceleration measurements collected by smartphone sensors during vehicle travel over different road surfaces. Originally developed for pavement type classification, we adapt it for irregular time series forecasting. The processed dataset contains 2,109 instances with 3 variables. We use the previous 400 steps to predict the acceleration measurements over the subsequent 50/100/200 steps.

\textbf{FNSPID}~\citep{DBLP:conf/kdd/DongFP24}. The FNSPID dataset is a comprehensive Financial News and Stock Price Integration Dataset, combining stock price movements with corresponding financial news headlines. It covers thousands of stocks from major exchanges, featuring irregularly timed news events alongside stock price variables such as open, close, and volume. The alignment of sparse news signals with continuous price data results in multiple multi-modal IMTS. We select 10 IMTS and 6 variables for irregular time series forecasting. To facilitate stock trend and volatility forecasting, each instance is organized into windows of sequential market activities. We use the integrated data from the previous 730 days as observed data to predict the stock price fluctuations and market sentiment for the following 30/180/365 days.

\textbf{GDELT}~\citep{leetaru2013gdelt}. This dataset is sourced from the Global Database of Events, Language, and Tone, which monitors world broadcast, print, and web news in multiple languages. It captures physical activities and socio-political events globally, resulting in a massive collection of IMTS with varying temporal resolutions based on event occurrences. The dataset includes multiple variables such as the Goldstein scale, average tone, and event counts across thousands of distinct themes and locations. To adapt it for irregular time series forecasting, we select 6 IMTS instances and 4 variables. We use the event dynamics from the previous 365 days as the observed data to predict the event intensity and regional stability in the subsequent 30/90/180 days.

\textbf{Seabirds}~\citep{browning2018predicting}. This dataset tracks the movement and behavioral patterns of seabirds using high-resolution GPS and tri-axial accelerometer sensors. It contains IMTS representing multiple distinct movement variables, such as altitude, speed, and body orientation, collected from multiple species during foraging trips. Given the irregular nature of wildlife tracking due to varying signal quality and nesting cycles, the data is segmented into multiple IMTS instances, each spanning a specific duration of flight or diving activity. We select 108 IMTS and 3 variables for irregular time series forecasting. We use the sensor measurements from the first 1440 minutes as the observed data to predict the bird's trajectory and behavioral states in the subsequent 60/360/720 minutes.

\begin{table}[h]
\centering
\caption{The statistics of evaluation datasets.}
\label{tab:dataset_stats}

\footnotesize
\setlength{\tabcolsep}{2.5pt}
\renewcommand{\arraystretch}{1.05}

\begin{adjustbox}{max width=\linewidth}
\begin{tabular}{@{}llcccccccc@{}}
\toprule
Dataset
& Domain
& \# Variables
& \# Samples
& \makecell{Avg. \\ \# Obs.}
& \makecell{Max\\Length}
& \makecell{Missing\\Rate}
& \makecell{Missing Pattern\\Complexity}
& \makecell{Sampling\\Irregularity}
& Skewness \\
\midrule
PhysioNet     & Healthcare    & 41 & 12000 & 74    & 48.00     & 0.856 & 0.614 & 0.640 & 0.180  \\
Human Activity & Activity      & 12 & 25    & 6593  & 304861.22 & 0.750 & 0.165 & 0.396 & 0.116  \\
USHCN         & Climate       & 5  & 1114  & 314   & 48.00     & 0.780 & 0.567 & 0.999 & 2.001  \\
CESNET        & Cybersecurity & 5  & 11    & 1700  & 1999.87   & 0.000 & 0.000 & 0.637 & 1.497  \\
Pamap2        & Healthcare    & 4  & 2     & 40601 & 540880.10 & 0.229 & 0.122 & 0.005 & -0.032 \\
EPA-Air       & Environment   & 4  & 8     & 5273  & 6575.00   & 0.708 & 0.171 & 0.057 & 1.505  \\
ClusterTrace  & Engineering   & 3  & 3     & 5005  & 1525.58   & 0.057 & 0.119 & 4.111 & 0.795  \\
APTC          & Engineering   & 3  & 2109  & 399   & 1232.00   & 0.001 & 0.001 & 0.000 & -0.060 \\
FNSPID        & Finance       & 6  & 10    & 3495  & 5303.00   & 0.001 & 0.001 & 0.606 & 1.100  \\
GDELT         & Social        & 4  & 6     & 1684  & 1460.69   & 0.042 & 0.188 & 1.279 & 3.140  \\
Seabirds      & Nature        & 3  & 108   & 2442  & 10078.00  & 0.001 & 0.001 & 0.283 & 0.691  \\
\bottomrule
\end{tabular}
\end{adjustbox}

\end{table}

\subsection{Metrics}
\label{appendix_metrics}

The metrics are defined as follows.
\begin{equation}
\begin{aligned}
\mathit{MSE}
&= \frac{1}{n}\sum_{i=1}^{n}(y_i-\hat{y}_i)^2,
\qquad
\mathit{MAE}
= \frac{1}{n}\sum_{i=1}^{n}|y_i-\hat{y}_i|,
\\[3pt]
\mathit{MAPE}
&= \frac{1}{n}\sum_{i=1}^{n}
\frac{|y_i-\hat{y}_i|}{|y_i|}\times 100\%,
\qquad
\mathit{SMAPE}
= \frac{100\%}{n}\sum_{i=1}^{n}
\frac{|\hat{y}_i-y_i|}
{(|y_i|+|\hat{y}_i|)/2},
\\[3pt]
\mathit{RMSE}
&= \sqrt{\frac{1}{n}\sum_{i=1}^{n}
(\hat{y}_i-y_i)^2},
\qquad
\mathit{WAPE}
= \frac{\sum_{i=1}^{n}|y_i-\hat{y}_i|}
{\sum_{i=1}^{n}|y_i|},
\\[3pt]
\mathit{MSMAPE}
&= \frac{100\%}{n}\sum_{i=1}^{n}
\frac{|\hat{y}_i-y_i|}
{\max(|y_i|+|\hat{y}_i|+\epsilon,\,0.5+\epsilon)/2},
\\[3pt]
\mathit{PCC}
&= \frac{
\sum_{i=1}^{n}(y_i-\bar{y})(\hat{y}_i-\bar{\hat{y}})
}{
\sqrt{\sum_{i=1}^{n}(y_i-\bar{y})^2}
\sqrt{\sum_{i=1}^{n}(\hat{y}_i-\bar{\hat{y}})^2}
},
\\[3pt]
R^2
&= 1-
\frac{\sum_{i=1}^{n}(y_i-\hat{y}_i)^2}
{\sum_{i=1}^{n}(y_i-\bar{y})^2},
\qquad
\mathit{CPN}
= |\mathit{CP}(\hat{\mathbf{y}})
-\mathit{CP}(\mathbf{y})|,
\\[3pt]
\mathit{DA}
&= \frac{1}{n-1}\sum_{i=1}^{n-1}
\mathbb{I}\!\left[
\mathit{sign}(y_{i+1}-y_i)
=
\mathit{sign}(\hat{y}_{i+1}-\hat{y}_i)
\right],
\\[3pt]
\mathit{DTW}
&= \min_{\pi}\sum_{(i,j)\in\pi}
\|\hat{y}_i-y_j\|,
\qquad
\mathit{IER}
= \frac{H(\hat{\mathbf{y}})}{H(\mathbf{y})}.
\end{aligned}
\end{equation}
where $y_i$ represents the actual value, $\hat{y}_i$ represents the predicted value, n is the number of observations, $sign(\cdot)$ is the sign function, and $\mathbb{I}[\cdot]$ is the indicator function which takes the value 1 if the condition inside the brackets is met, and 0 otherwise. $\pi$ denotes a valid alignment path. $H(\cdot)$ denotes the information entropy of the time series. $\mathit{CP}(\mathbf{y}) = \sum_{i=3}^{n} \mathbb{I} \left( \mathit{sign}(\Delta y_i) \neq \text{sign}(\Delta y_{i-1}) \right)$. We set the parameter $\epsilon$ to its proposed default of 0.1. For error-based metrics, lower values indicate better performance. Below we introduce the non-error-based metrics. 

\begin{itemize}
    \item Pearson Correlation (PCC) measures the linear relationship between the predicted series and the ground-truth series. Its value ranges from $-1$ to $1$, where higher values indicate that the model better captures the trend of the time series. A value of 0 indicates no linear correlation, while negative values indicate an inverse trend.
    
    \item Coefficient of Determination ($R^2$) measures how well the model’s predictions account for the variance in the true series. Its value ranges from $-\infty$ to $1$, where values closer to $1$ indicate that the model better captures the variability of the true data. A negative value suggests that the model performs worse than a simple mean-based prediction.
    
    \item Directional Accuracy (DA) measures whether the model correctly predicts the direction of change in the time series. Its value ranges from 0 to 1, with higher values indicating a better ability to capture the trend direction.
    \item Dynamic Time Warping (DTW) measures the overall similarity between the predicted series and the ground-truth series under a non-linear alignment of the time axis. Lower values indicate that the predicted series is more similar to the ground-truth series in terms of overall shape, reflecting better forecasting performance.
    
    \item Change Point Number Difference (CPN) measures the difference in the number of change points between the predicted series and the ground-truth series. Smaller values indicate that the predicted series is closer to the ground-truth in terms of structural changes and periodic patterns, reflecting better forecasting performance.
    
    \item Information Entropy Ratio (IER) measures the consistency between the predicted series and the ground-truth series in terms of complexity and uncertainty. Values closer to 1 indicate that the predicted series matches the ground-truth in terms of complexity and information content, reflecting better forecasting performance. Values significantly less than 1 suggest that the predicted series is overly smooth, while values significantly greater than 1 indicate that the prediction contains excessive noise.
\end{itemize}

\subsection{Irregular Time Series Characteristics}
\label{data_characteristics_appendix}

\textbf{Missing Rate.} Missing Rate measures the proportion of missing values in the dataset based on the original observation mask, reflecting the overall degree of data incompleteness. Algorithm~\ref{alg:missing_rate} details the calculation process.

\begin{algorithm}[H]
\caption{Calculating Missing Rate}
\label{alg:missing_rate}
\begin{algorithmic}[1]
\renewcommand{\algorithmicrequire}{\textbf{Input:}}
\renewcommand{\algorithmicensure}{\textbf{Output:}}
\REQUIRE Mask matrix $M \in \{0, 1\}^{T \times D}$ (1 for observed, 0 for missing)
\ENSURE Missing rate value $\rho \in [0, 1]$
\STATE $T, D \leftarrow \text{dimensions of } M$
\STATE $total\_elements \leftarrow T \times D$
\STATE $missing\_count \leftarrow \sum_{i=1}^{T} \sum_{j=1}^{D} (1 - M_{ij})$
\STATE \textbf{return} $\rho \leftarrow \frac{missing\_count}{total\_elements}$
\end{algorithmic}
\end{algorithm}

\textbf{Missing Pattern Complexity.}
Missing Pattern Complexity measures the structural complexity and uncertainty of missing patterns across different time steps based on the original observation mask. We use missing entropy to represent the missing pattern complexity. Algorithm~\ref{alg:missing_entropy} details the calculation process.

\begin{algorithm}[H]
\caption{Calculating Missing Entropy}
\label{alg:missing_entropy}
\begin{algorithmic}[1]
\renewcommand{\algorithmicrequire}{\textbf{Input:}}
\renewcommand{\algorithmicensure}{\textbf{Output:}}
\REQUIRE Mask matrix $M \in \{0, 1\}^{T \times D}$
\ENSURE Normalized missing entropy $\eta \in [0, 1]$
\STATE Define missing pattern vectors $\mathbf{r}_i = (1-M_{i1}, \dots, 1-M_{iD}) \in \{0,1\}^D$ for $i=1 \dots T$
\STATE Identify $K$ unique patterns $\{\boldsymbol{\pi}_1, \dots, \boldsymbol{\pi}_K\}$ from $\{\mathbf{r}_1, \dots, \mathbf{r}_T\}$
\FOR{each unique pattern $\boldsymbol{\pi}_k$}
\STATE $p_k \leftarrow \frac{1}{T} \sum_{i=1}^{T} 1(\mathbf{r}_i = \boldsymbol{\pi}_k)$
\ENDFOR
\STATE $H \leftarrow -\sum_{k=1}^{K} p_k \log(p_k)$
\STATE $H_{max} \leftarrow \min(\log T, D \log 2)$
\STATE \textbf{return} $\eta \leftarrow \frac{H}{H_{max} + \epsilon}$ \COMMENT{where $\epsilon = 10^{-12}$}
\end{algorithmic}
\end{algorithm}

\textbf{Sampling Irregularity.}
Sampling Irregularity measures the irregularity of sampling times using the coefficient of variation of adjacent sampling intervals. We use coefficient of variation of adjacent sampling intervals to represent the sampling irregularity. Algorithm~\ref{alg:time_cv} details the calculation process.

\begin{algorithm}[H]
\caption{Calculating Sampling Irregularity}
\label{alg:time_cv}
\begin{algorithmic}[1]
\renewcommand{\algorithmicrequire}{\textbf{Input:}}
\renewcommand{\algorithmicensure}{\textbf{Output:}}
\REQUIRE Observation time stamps $t = \{t_1, t_2, \dots, t_T\}$
\ENSURE Coefficient of variation for intervals $CV_{time}$
\STATE Sort $t$ such that $t_{(1)} \le t_{(2)} \le \dots \le t_{(T)}$
\FOR{$i = 1$ \textbf{to} $T-1$}
\STATE $\Delta_i \leftarrow t_{(i+1)} - t_{(i)}$
\ENDFOR
\STATE $\mu_{\Delta} \leftarrow \text{mean}(\Delta_1, \dots, \Delta_{T-1})$
\STATE $\sigma_{\Delta} \leftarrow \text{std}(\Delta_1, \dots, \Delta_{T-1})$
\STATE \textbf{return} $CV_{time} \leftarrow \frac{\sigma_{\Delta}}{\mu_{\Delta} + \epsilon}$
\end{algorithmic}
\end{algorithm}

\textbf{Skewness.}
Skewness measures the overall asymmetry of the data distribution by averaging the skewness values computed separately for each variable. Algorithm~\ref{alg:skewness} details the calculation process.

\begin{algorithm}[H]
\caption{Calculating Average Skewness}
\label{alg:skewness}
\begin{algorithmic}[1]
\renewcommand{\algorithmicrequire}{\textbf{Input:}}
\renewcommand{\algorithmicensure}{\textbf{Output:}}
\REQUIRE Time series $X \in \mathbb{R}^{T \times D}$, Mask matrix $M \in \{0, 1\}^{T \times D}$
\ENSURE Average skewness value $\gamma_{avg}$
\STATE $\mathcal{J} \leftarrow \emptyset$ \COMMENT{Set of valid variables}
\FOR{each variable $j = 1$ \textbf{to} $D$}
\STATE Extract observed values $x^{(j)} = \{X_{ij} \mid M_{ij} = 1\}$
\IF{$|x^{(j)}| > \text{threshold}$}
\STATE $\mu_j, \sigma_j \leftarrow \text{mean}(x^{(j)}), \text{std}(x^{(j)})$
\STATE $\gamma_j \leftarrow \frac{\frac{1}{n_j} \sum_{k=1}^{n_j} (x_k^{(j)} - \mu_j)^3}{\sigma_j^3}$
\STATE $\mathcal{J} \leftarrow \mathcal{J} \cup \{j\}$
\ENDIF
\ENDFOR
\STATE \textbf{return} $\gamma_{avg} \leftarrow \frac{1}{|\mathcal{J}|} \sum_{j \in \mathcal{J}} \gamma_j$
\end{algorithmic}
\end{algorithm}

\subsection{Other Experiment results}
\label{sec: other results appendix}

\subsubsection{Model Efficiency Analysis}
\label{sec: Model Efficiency Analysis}

\begin{figure}[h]
    \centering
    \includegraphics[width=\linewidth]{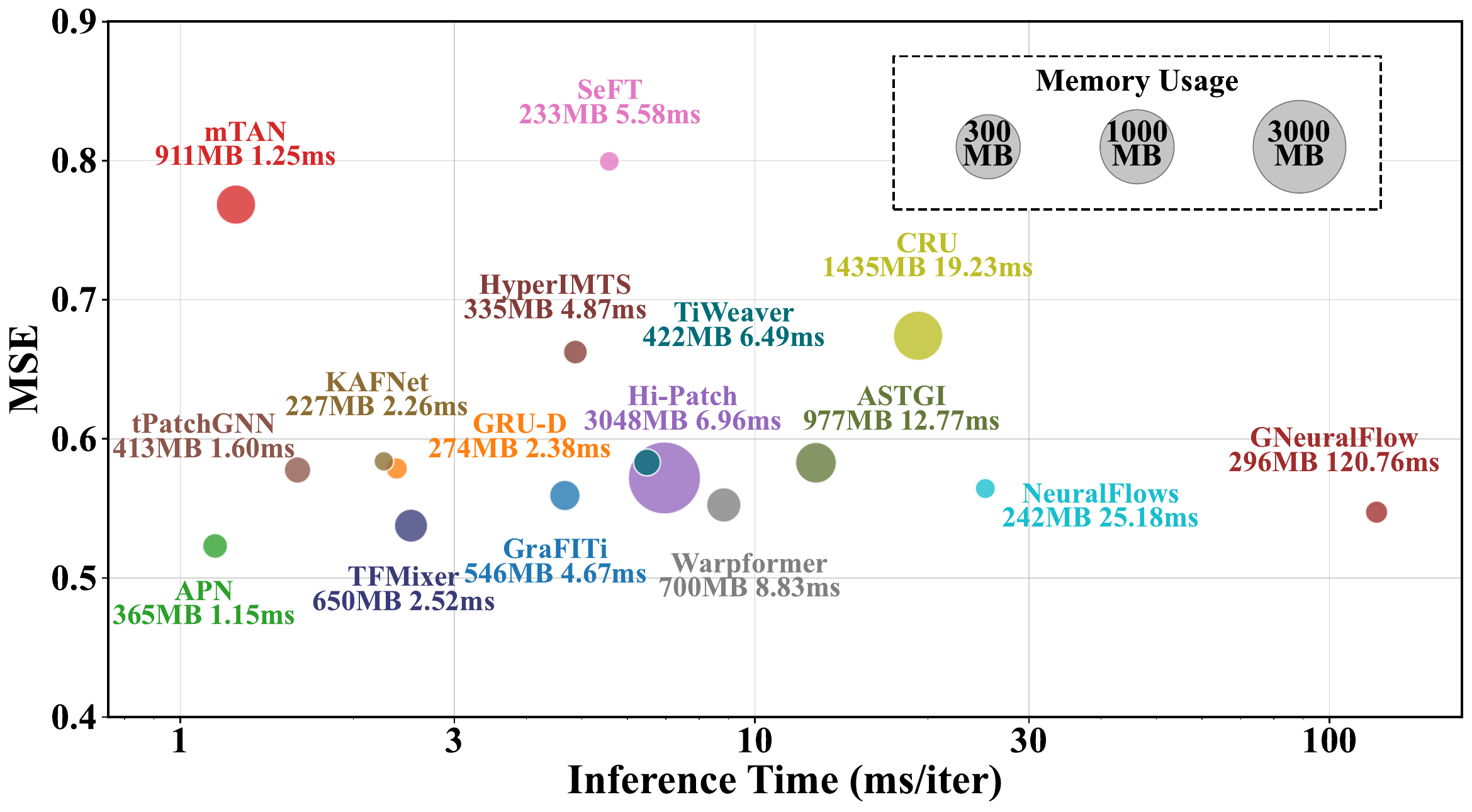}
    \caption{Inference efficiency comparison under BITS.}
    \label{fig:efficiency}
\end{figure}

We compare six categories of irregular forecasting methods on USHCN in terms of MSE, inference time, and memory usage, as shown in Figure~\ref{fig:efficiency}. Among temporal segmentation and aggregation methods, APN achieves the lowest MSE and fastest inference with moderate memory consumption, demonstrating that effective temporal aggregation does not necessarily require substantial computational resources. However, efficiency varies considerably within this category. TiWeaver consumes less memory than TFMixer but incurs higher inference latency, while Hi-Patch requires substantially more memory, potentially due to the additional overhead of hierarchical patch-graph modeling. A similar disparity appears among data-format transformation methods, where ASTGI incurs higher latency than GraFITi and HyperIMTS, possibly reflecting the computational cost of adaptive spatio-temporal graph construction. Differential equation-based methods generally incur longer inference times, particularly GNeuralFlow, which suggests that modeling continuous dynamics can introduce substantial computational overhead. In contrast, imputation-based mTAN achieves low inference latency but requires relatively high memory, while sampling context-aware GRU-D maintains low memory consumption through relatively lightweight recurrent modeling. Multi-scale Warpformer also incurs additional latency, potentially associated with temporal warping and multi-scale representation learning. These results highlight that greater modeling complexity does not necessarily yield better accuracy, and low memory consumption does not guarantee fast inference. Therefore, forecasting accuracy, inference latency, and memory usage should be jointly considered when evaluating irregular forecasting methods.

\subsubsection{Full results of performance comparison on irregular time series forecasting}
\label{sec: full results}

We provide the full results in Table~\ref{tab:mse_results}, Table~\ref{tab:mae_results}, Table~\ref{tab:regular_tsfm_mse_results} and Table~\ref{tab:regular_tsfm_mae_results}.

\begin{table}[h]
\centering
\caption{Performance comparison of irregular time series forecasting models in terms of MSE ($\times 10^{-2}$). Best results are in \textbf{bold} and second-best are \underline{underlined}.}
\label{tab:mse_results}

\setlength{\tabcolsep}{2.0pt}
\renewcommand{\arraystretch}{1.08}

\makebox[\textwidth][c]{%
\resizebox{\textwidth}{!}{%
\begin{tabular}{c|c|cccccccccccccccc}
\toprule
\textbf{Dataset} & \textbf{Pred} &
\textbf{TFM} &
\textbf{TiW} &
\textbf{AST} &
\textbf{APN} &
\textbf{KAF} &
\textbf{HiP} &
\textbf{HIMTS} &
\textbf{GNF} &
\textbf{tPGNN} &
\textbf{GFT} &
\textbf{Warp} &
\textbf{CRU} &
\textbf{NF} &
\textbf{mTAN} &
\textbf{SeFT} &
\textbf{GRU-D} \\
\midrule

\multirow{4}{*}{PhysioNet}
& 1
& 0.619 & 0.606 & 0.553 & \textbf{0.439} & 0.654 & 0.559
& 0.493 & 1.220 & \underline{0.455} & 0.594 & 1.290
& 1.298 & 1.136 & 1.029 & 1.120 & 1.082 \\

& 6
& 0.486 & 0.522 & 0.470 & \textbf{0.443} & 0.497 & 0.528
& 0.481 & 0.929 & \underline{0.443} & 0.444 & 1.018
& 1.114 & 0.943 & 0.798 & 1.160 & 0.909 \\

& 12
& 0.569 & 0.573 & \underline{0.530} & 0.530 & 0.550 & 0.566
& 0.791 & 0.972 & \textbf{0.507} & 0.663 & 1.066
& 1.088 & 0.909 & 0.842 & 1.251 & 0.879 \\

\cmidrule(l){2-18}
& \textbf{Avg}
& 0.558 & 0.567 & 0.518 & \underline{0.471} & 0.567 & 0.551
& 0.588 & 1.040 & \textbf{0.468} & 0.567 & 1.125
& 1.167 & 0.996 & 0.890 & 1.177 & 0.957 \\
\midrule

\multirow{4}{*}{Human Activity}
& 200
& \textbf{0.254} & 0.278 & 0.277 & \underline{0.259} & 0.265 & 0.292
& 0.261 & 1.852 & 0.266 & 0.259 & 0.396
& 0.819 & 0.977 & 0.521 & 0.841 & 0.399 \\

& 1200
& \textbf{0.354} & 0.390 & 0.385 & \underline{0.358} & 0.373 & 0.401
& 0.369 & 1.877 & 0.403 & 0.373 & 1.704
& 0.871 & 1.017 & 0.540 & 1.691 & 0.714 \\

& 3000
& 0.481 & 0.538 & 0.509 & \textbf{0.481} & 0.511 & 0.492
& 0.542 & 1.794 & \underline{0.481} & 0.502 & 2.079
& 1.001 & 1.019 & 0.701 & 1.675 & 0.569 \\

\cmidrule(l){2-18}
& \textbf{Avg}
& \textbf{0.363} & 0.402 & 0.390 & \underline{0.366} & 0.383 & 0.395
& 0.391 & 1.841 & 0.383 & 0.378 & 1.393
& 0.897 & 1.004 & 0.587 & 1.402 & 0.561 \\
\midrule

\multirow{4}{*}{USHCN}
& 1
& \underline{53.76} & 58.31 & 58.28 & \textbf{52.30} & 58.38 & 57.19
& 66.24 & 54.73 & 57.76 & 55.93 & 55.25
& 67.41 & 56.43 & 76.84 & 79.95 & 57.87 \\

& 6
& \textbf{48.50} & 51.16 & \underline{49.03} & 60.86 & 65.30 & 63.62
& 68.48 & 51.62 & 74.51 & 50.98 & 50.58
& 61.28 & 57.50 & 68.92 & 74.17 & 74.00 \\

& 12
& \textbf{49.18} & 58.77 & \underline{49.78} & 51.42 & 69.73 & 54.16
& 72.17 & 71.03 & 75.36 & 71.83 & 53.74
& 61.77 & 57.86 & 58.70 & 75.12 & 75.01 \\

\cmidrule(l){2-18}
& \textbf{Avg}
& \textbf{50.48} & 56.08 & \underline{52.36} & 54.86 & 64.47 & 58.32
& 68.96 & 59.13 & 69.21 & 59.58 & 53.19
& 63.49 & 57.26 & 68.15 & 76.41 & 68.96 \\
\midrule

\multirow{4}{*}{CESNET}
& 24
& 2.293 & \textbf{2.153} & \underline{2.164} & 2.298 & 2.336 & 2.340
& 2.297 & 2.323 & 2.313 & 2.234 & 2.711
& 2.308 & 2.304 & 2.329 & 3.028 & 2.617 \\

& 168
& 2.332 & \textbf{2.156} & \underline{2.160} & 2.307 & 2.362 & 2.443
& 2.696 & 2.343 & 2.339 & 2.307 & 2.339
& 2.331 & 2.348 & 2.348 & 5.077 & 2.326 \\

& 336
& 2.357 & \underline{2.160} & \textbf{2.156} & 2.353 & 2.355 & 2.523
& 2.978 & 2.351 & 2.351 & 2.303 & 4.098
& 2.376 & 2.340 & 2.324 & 5.082 & 2.328 \\

\cmidrule(l){2-18}
& \textbf{Avg}
& 2.327 & \textbf{2.156} & \underline{2.160} & 2.319 & 2.351 & 2.435
& 2.657 & 2.339 & 2.334 & 2.281 & 3.049
& 2.338 & 2.331 & 2.334 & 4.396 & 2.424 \\
\midrule

\multirow{4}{*}{Pamap2}
& 200
& 0.294 & 0.274 & 0.246 & \underline{0.225} & 0.305 & 0.365
& 0.308 & 1.619 & 0.335 & \textbf{0.216} & -
& 0.847 & 1.540 & 0.350 & 1.860 & 1.770 \\

& 1200
& \underline{0.635} & 0.667 & 0.651 & 0.754 & 0.830 & 0.722
& \textbf{0.609} & 1.796 & 0.757 & 0.697 & -
& 1.574 & 1.630 & 1.140 & 1.881 & 1.803 \\

& 3000
& \textbf{0.869} & \underline{0.943} & 1.042 & 1.914 & 1.879 & 1.495
& 0.946 & 1.996 & 1.465 & 1.150 & -
& 1.470 & 1.480 & 1.127 & 1.880 & 1.752 \\

\cmidrule(l){2-18}
& \textbf{Avg}
& \textbf{0.599} & 0.628 & 0.646 & 0.964 & 1.005 & 0.861
& \underline{0.621} & 1.804 & 0.852 & 0.688 & -
& 1.297 & 1.550 & 0.872 & 1.874 & 1.775 \\
\midrule

\multirow{4}{*}{EPA-Air}
& 24
& \textbf{0.742} & 0.807 & \underline{0.790} & 2.093 & 1.613 & 1.104
& 0.851 & 2.530 & 1.749 & 0.791 & 2.001
& 1.623 & 1.747 & 2.076 & 2.949 & 2.401 \\

& 72
& 0.881 & \underline{0.830} & \textbf{0.766} & 1.454 & 1.567 & 1.146
& 0.841 & 2.241 & 1.587 & 0.946 & 2.461
& 1.563 & 1.979 & 2.027 & 2.718 & 1.546 \\

& 168
& \textbf{0.750} & 0.803 & \underline{0.778} & 1.647 & 1.438 & 0.926
& 1.336 & 1.937 & 1.400 & 0.778 & 2.341
& 1.330 & 1.474 & 1.883 & 2.414 & 2.202 \\

\cmidrule(l){2-18}
& \textbf{Avg}
& \underline{0.791} & 0.813 & \textbf{0.778} & 1.731 & 1.539 & 1.059
& 1.009 & 2.236 & 1.579 & 0.838 & 2.268
& 1.505 & 1.733 & 1.995 & 2.694 & 2.050 \\
\midrule

\multirow{4}{*}{ClusterTrace}
& 24
& \textbf{3.919} & \underline{3.997} & 4.060 & 4.250 & 4.497 & 4.372
& 4.470 & 4.721 & 4.619 & 4.478 & 4.600
& 4.853 & 4.471 & 5.058 & 4.950 & 4.630 \\

& 48
& 4.356 & 4.344 & \textbf{4.249} & 4.465 & 4.485 & \underline{4.285}
& 4.316 & 4.471 & 4.463 & 4.459 & 4.710
& 4.797 & 4.471 & 4.923 & 4.980 & 4.464 \\

& 72
& 4.321 & 4.656 & \textbf{4.222} & \underline{4.237} & 4.495 & 4.247
& 4.432 & 4.536 & 4.256 & 4.557 & 4.571
& 5.161 & 4.609 & 4.316 & 4.800 & 4.418 \\

\cmidrule(l){2-18}
& \textbf{Avg}
& \underline{4.199} & 4.332 & \textbf{4.177} & 4.317 & 4.492 & 4.301
& 4.406 & 4.576 & 4.446 & 4.498 & 4.627
& 4.937 & 4.517 & 4.766 & 4.910 & 4.504 \\
\midrule

\multirow{4}{*}{APTC}
& 50
& 0.391 & 0.391 & \textbf{0.390} & \underline{0.390} & 0.391 & 0.392
& 0.391 & 0.395 & 0.561 & 0.391 & 0.571
& 0.391 & 0.394 & 0.392 & 0.430 & 0.391 \\

& 100
& 0.377 & 0.377 & 0.383 & 0.379 & \textbf{0.377} & 0.377
& \underline{0.377} & 0.378 & 0.624 & 0.377 & 0.634
& 0.377 & 0.379 & 0.377 & 0.436 & 0.377 \\

& 200
& 0.376 & \textbf{0.376} & 0.376 & 0.377 & \underline{0.376} & 0.376
& 0.376 & 0.376 & 0.622 & 0.376 & 0.401
& 0.376 & 0.379 & 0.376 & 0.412 & 0.376 \\

\cmidrule(l){2-18}
& \textbf{Avg}
& \textbf{0.381} & \underline{0.381} & 0.383 & 0.382 & 0.381 & 0.382
& 0.381 & 0.383 & 0.602 & 0.381 & 0.535
& 0.381 & 0.384 & 0.382 & 0.426 & 0.381 \\
\midrule

\multirow{4}{*}{FNSPID}
& 30
& \underline{0.157} & \textbf{0.155} & 1.490 & 0.606 & 0.373 & 0.313
& 0.839 & 0.831 & 0.848 & 0.527 & 1.940
& 1.311 & 0.805 & 0.289 & 1.826 & 0.359 \\

& 180
& \textbf{0.202} & \underline{0.232} & 0.717 & 0.318 & 0.350 & 0.661
& 0.555 & 0.910 & 0.645 & 0.663 & 1.771
& 1.011 & 0.764 & 0.888 & 1.690 & 1.289 \\

& 365
& 0.638 & \underline{0.514} & 1.001 & 1.431 & \textbf{0.396} & 1.011
& 0.528 & 1.073 & 1.187 & 1.464 & 1.691
& 1.073 & 0.612 & 0.802 & 1.651 & 1.877 \\

\cmidrule(l){2-18}
& \textbf{Avg}
& \underline{0.332} & \textbf{0.300} & 1.069 & 0.785 & 0.373 & 0.662
& 0.641 & 0.938 & 0.893 & 0.885 & 1.801
& 1.132 & 0.727 & 0.660 & 1.722 & 1.175 \\
\midrule

\multirow{4}{*}{GDELT}
& 30
& 1.913 & 1.899 & \textbf{1.876} & 1.902 & 1.893 & 1.945
& 1.882 & 1.900 & 1.912 & \underline{1.878} & 2.100
& 1.897 & 1.902 & 2.047 & 2.880 & 1.891 \\

& 90
& 1.918 & \textbf{1.874} & \underline{1.884} & 1.902 & 1.886 & 1.902
& 1.914 & 1.912 & 1.920 & 2.105 & 2.690
& 1.964 & 1.902 & 1.921 & 2.650 & 1.889 \\

& 180
& 1.746 & \textbf{1.713} & \underline{1.719} & 1.719 & 1.732 & 2.018
& 1.751 & 1.782 & 1.746 & 1.876 & 2.130
& 1.730 & 1.724 & 1.786 & 2.631 & 1.731 \\

\cmidrule(l){2-18}
& \textbf{Avg}
& 1.859 & \underline{1.829} & \textbf{1.826} & 1.841 & 1.837 & 1.955
& 1.849 & 1.865 & 1.859 & 1.953 & 2.307
& 1.864 & 1.843 & 1.918 & 2.720 & 1.837 \\
\midrule

\multirow{4}{*}{Seabirds}
& 60
& \underline{0.016} & 0.018 & \textbf{0.008} & 0.036 & 0.083 & 0.103
& 0.063 & 0.111 & 0.116 & 0.110 & 0.081
& 0.063 & 0.197 & 0.131 & 0.121 & 0.141 \\

& 360
& 0.034 & \underline{0.020} & \textbf{0.020} & 0.067 & 0.094 & 0.061
& 0.048 & 0.131 & 0.098 & 0.047 & 0.120
& 0.130 & 0.173 & 0.150 & 0.140 & 0.181 \\

& 720
& 0.046 & 0.040 & \underline{0.039} & 0.139 & 0.114 & 0.130
& \textbf{0.035} & 0.161 & 0.138 & 0.067 & 0.150
& 0.150 & 0.140 & 0.068 & 0.161 & 0.170 \\

\cmidrule(l){2-18}
& \textbf{Avg}
& 0.032 & \underline{0.026} & \textbf{0.022} & 0.081 & 0.097 & 0.098
& 0.049 & 0.134 & 0.117 & 0.075 & 0.117
& 0.114 & 0.170 & 0.116 & 0.141 & 0.164 \\

\bottomrule
\end{tabular}%
}%
}

\vspace{1pt}
\parbox{\textwidth}{%
\scriptsize
\raggedright
\textit{Note.} ``-'' indicates out-of-memory (OOM) even with the batch size set to 1. Avg is computed from the three displayed prediction-length values.
\textit{Abbreviations.}
TFM: TFMixer;
TiW: TiWeaver;
AST: ASTGI;
KAF: KAFNet;
HiP: Hi-Patch;
HIMTS: HyperIMTS;
GNF: GNeuralFlow;
tPGNN: tPatchGNN;
GFT: GraFITi;
Warp: Warpformer;
NF: NeuralFlows.
}

\end{table}

\begin{table}[h]
\centering
\caption{Performance comparison of irregular time series forecasting models in terms of MAE ($\times 10^{-2}$). Best results are in \textbf{bold} and second-best are \underline{underlined}.}
\label{tab:mae_results}

\setlength{\tabcolsep}{1.2pt}
\renewcommand{\arraystretch}{1.08}

\makebox[\textwidth][c]{%
\resizebox{\textwidth}{!}{%
\begin{tabular}{c|c|cccccccccccccccc}
\toprule
\textbf{Dataset} & \textbf{Pred} &
\textbf{TFM} &
\textbf{TiW} &
\textbf{AST} &
\textbf{APN} &
\textbf{KAF} &
\textbf{HiP} &
\textbf{HIMTS} &
\textbf{GNF} &
\textbf{tPGNN} &
\textbf{GFT} &
\textbf{Warp} &
\textbf{CRU} &
\textbf{NF} &
\textbf{mTAN} &
\textbf{SeFT} &
\textbf{GRU-D} \\
\midrule

\multirow{4}{*}{PhysioNet}
& 1
& 3.765 & 3.555 & 3.882 & \underline{3.554} & 4.980 & 4.004
& \textbf{3.454} & 6.061 & 3.706 & 3.972 & 6.646
& 6.366 & 5.898 & 5.576 & 5.891 & 5.524 \\

& 6
& \textbf{3.415} & 3.636 & 3.692 & 3.659 & 3.763 & 3.963
& 3.914 & 5.548 & \underline{3.540} & 3.604 & 5.250
& 5.891 & 5.609 & 5.093 & 5.391 & 5.525 \\

& 12
& \textbf{3.525} & 3.729 & 3.805 & 3.751 & 4.110 & 3.859
& 5.451 & 5.869 & \underline{3.710} & 4.538 & 5.453
& 6.016 & 5.298 & 5.052 & 5.281 & 5.144 \\

\cmidrule(l){2-18}
& \textbf{Avg}
& \textbf{3.568} & \underline{3.640} & 3.793 & 3.655 & 4.284 & 3.942
& 4.273 & 5.826 & 3.652 & 4.038 & 5.783
& 6.091 & 5.602 & 5.240 & 5.521 & 5.398 \\
\midrule

\multirow{4}{*}{Human Activity}
& 200
& \textbf{2.814} & 3.060 & 3.167 & 2.974 & 3.054 & 3.373
& 3.028 & 10.641 & 3.029 & \underline{2.963} & 4.365
& 6.450 & 7.523 & 5.191 & 6.190 & 4.341 \\

& 1200
& \textbf{3.568} & 3.875 & 4.001 & \underline{3.688} & 3.871 & 4.149
& 3.831 & 10.459 & 4.156 & 3.949 & 10.427
& 6.724 & 7.806 & 5.310 & 10.294 & 6.300 \\

& 3000
& \textbf{4.431} & 4.794 & 4.843 & \underline{4.534} & 4.826 & 4.859
& 5.153 & 10.411 & 4.718 & 4.794 & 11.317
& 7.343 & 7.725 & 6.190 & 10.249 & 5.504 \\

\cmidrule(l){2-18}
& \textbf{Avg}
& \textbf{3.604} & 3.910 & 4.004 & \underline{3.732} & 3.917 & 4.127
& 4.004 & 10.504 & 3.968 & 3.902 & 8.703
& 6.839 & 7.685 & 5.564 & 8.911 & 5.382 \\
\midrule

\multirow{4}{*}{USHCN}
& 1
& \textbf{25.15} & 33.81 & 37.09 & \underline{30.97} & 35.16 & 36.97
& 50.59 & 35.49 & 36.39 & 33.34 & 40.90
& 42.81 & 35.06 & 45.15 & 50.37 & 36.52 \\

& 6
& \textbf{24.40} & \underline{31.15} & 32.70 & 41.02 & 43.41 & 43.73
& 43.97 & 34.17 & 50.86 & 33.25 & 35.67
& 41.05 & 37.23 & 44.83 & 50.18 & 49.05 \\

& 12
& \textbf{27.79} & 35.54 & \underline{33.58} & 35.70 & 47.81 & 40.37
& 46.68 & 46.62 & 51.09 & 52.29 & 37.28
& 40.94 & 39.37 & 36.67 & 50.54 & 48.77 \\

\cmidrule(l){2-18}
& \textbf{Avg}
& \textbf{25.78} & \underline{33.50} & 34.46 & 35.90 & 42.13 & 40.36
& 47.08 & 38.76 & 46.11 & 39.63 & 37.95
& 41.60 & 37.22 & 42.22 & 50.36 & 44.78 \\
\midrule

\multirow{4}{*}{CESNET}
& 24
& 9.723 & \textbf{9.019} & \underline{9.384} & 10.211 & 10.363 & 10.174
& 10.050 & 10.262 & 10.371 & 9.981 & 11.545
& 10.316 & 10.243 & 10.382 & 11.363 & 11.318 \\

& 168
& 10.035 & \textbf{9.073} & \underline{9.209} & 10.166 & 10.369 & 10.365
& 12.280 & 10.392 & 10.272 & 10.424 & 10.162
& 10.378 & 10.400 & 10.383 & 15.399 & 10.305 \\

& 336
& 10.259 & \textbf{9.080} & \underline{9.428} & 10.101 & 10.414 & 10.436
& 13.340 & 10.358 & 10.352 & 10.389 & 13.600
& 10.687 & 10.288 & 10.367 & 15.442 & 10.333 \\

\cmidrule(l){2-18}
& \textbf{Avg}
& 10.006 & \textbf{9.057} & \underline{9.340} & 10.159 & 10.382 & 10.325
& 11.890 & 10.337 & 10.332 & 10.265 & 11.769
& 10.460 & 10.310 & 10.377 & 14.068 & 10.652 \\
\midrule

\multirow{4}{*}{Pamap2}
& 200
& 3.279 & 2.839 & 2.898 & \underline{2.762} & 3.201 & 4.386
& 3.448 & 9.100 & 4.063 & \textbf{2.649} & -
& 6.820 & 8.504 & 4.107 & 7.571 & 9.071 \\

& 1200
& 5.481 & \textbf{4.869} & 5.702 & 6.056 & 6.138 & 6.255
& \underline{5.253} & 9.644 & 6.142 & 5.954 & -
& 8.854 & 8.891 & 7.739 & 7.480 & 9.955 \\

& 3000
& \underline{6.387} & \textbf{5.885} & 7.556 & 10.277 & 10.803 & 8.859
& 6.578 & 10.076 & 9.481 & 7.570 & -
& 8.238 & 8.311 & 7.852 & 7.540 & 9.217 \\

\cmidrule(l){2-18}
& \textbf{Avg}
& \underline{5.049} & \textbf{4.531} & 5.385 & 6.365 & 6.714 & 6.500
& 5.093 & 9.607 & 6.562 & 5.391 & -
& 7.971 & 8.569 & 6.566 & 7.530 & 9.414 \\
\midrule

\multirow{4}{*}{EPA-Air}
& 24
& \underline{6.125} & \textbf{6.068} & 6.566 & 11.071 & 8.813 & 7.674
& 6.709 & 11.791 & 9.422 & 6.862 & 9.968
& 8.767 & 9.400 & 9.762 & 12.806 & 10.854 \\

& 72
& 6.632 & \textbf{6.220} & \underline{6.435} & 8.555 & 8.668 & 7.496
& 6.663 & 10.811 & 8.687 & 7.645 & 11.736
& 8.983 & 10.042 & 10.469 & 12.641 & 8.694 \\

& 168
& 6.419 & \underline{6.416} & \textbf{6.365} & 9.449 & 8.385 & 6.923
& 9.433 & 10.343 & 8.422 & 6.627 & 10.793
& 8.673 & 8.638 & 10.018 & 11.071 & 10.589 \\

\cmidrule(l){2-18}
& \textbf{Avg}
& \underline{6.392} & \textbf{6.235} & 6.455 & 9.692 & 8.622 & 7.364
& 7.602 & 10.982 & 8.844 & 7.045 & 10.832
& 8.808 & 9.360 & 10.083 & 12.173 & 10.046 \\
\midrule

\multirow{4}{*}{ClusterTrace}
& 24
& \textbf{14.730} & \underline{15.149} & 15.754 & 15.305 & 15.502 & 15.620
& 15.749 & 16.318 & 15.733 & 15.585 & 16.000
& 17.173 & 15.553 & 17.625 & 15.460 & 15.724 \\

& 48
& \underline{15.305} & 15.900 & \textbf{15.226} & 15.414 & 15.479 & 15.338
& 15.857 & 15.674 & 15.415 & 15.469 & 16.871
& 16.010 & 15.555 & 16.125 & 15.481 & 15.416 \\

& 72
& \textbf{15.137} & 16.400 & 16.051 & 16.036 & 15.422 & 16.036
& 15.627 & 15.840 & 15.853 & 15.701 & 17.101
& 16.336 & 15.626 & \underline{15.275} & 15.370 & 15.428 \\

\cmidrule(l){2-18}
& \textbf{Avg}
& \textbf{15.057} & 15.816 & 15.677 & 15.585 & 15.468 & 15.665
& 15.744 & 15.944 & 15.667 & 15.585 & 16.657
& 16.506 & 15.578 & 16.342 & \underline{15.437} & 15.523 \\
\midrule

\multirow{4}{*}{APTC}
& 50
& 4.679 & \textbf{4.677} & 4.683 & 4.693 & 4.678 & 4.684
& \underline{4.678} & 4.709 & 5.737 & 4.678 & 5.791
& 4.679 & 4.704 & 4.684 & 4.974 & 4.678 \\

& 100
& 4.580 & 4.580 & 4.629 & 4.594 & 4.580 & \textbf{4.580}
& 4.581 & 4.587 & 6.084 & \underline{4.580} & 6.150
& 4.581 & 4.601 & 4.585 & 5.084 & 4.580 \\

& 200
& 4.552 & 4.551 & 4.554 & 4.562 & 4.551 & \textbf{4.550}
& 4.552 & \underline{4.550} & 6.060 & 4.552 & 4.743
& 4.552 & 4.576 & 4.556 & 4.850 & 4.550 \\

\cmidrule(l){2-18}
& \textbf{Avg}
& 4.604 & \textbf{4.603} & 4.622 & 4.616 & 4.603 & 4.605
& 4.604 & 4.615 & 5.960 & 4.603 & 5.561
& 4.604 & 4.627 & 4.608 & 4.969 & \underline{4.603} \\
\midrule

\multirow{4}{*}{FNSPID}
& 30
& \underline{2.770} & \textbf{2.702} & 7.793 & 6.201 & 4.626 & 4.708
& 7.102 & 6.591 & 6.753 & 5.226 & 8.630
& 8.861 & 6.867 & 4.125 & 8.780 & 4.710 \\

& 180
& \underline{3.214} & \textbf{2.831} & 7.236 & 4.076 & 4.427 & 5.493
& 5.474 & 6.690 & 5.603 & 5.991 & 8.140
& 7.981 & 6.523 & 7.127 & 8.481 & 9.476 \\

& 365
& 5.466 & \textbf{4.152} & 8.451 & 9.210 & \underline{4.902} & 7.301
& 5.301 & 7.592 & 9.243 & 9.175 & 8.590
& 7.886 & 5.780 & 7.315 & 9.830 & 10.436 \\

\cmidrule(l){2-18}
& \textbf{Avg}
& \underline{3.817} & \textbf{3.228} & 7.827 & 6.496 & 4.652 & 5.834
& 5.959 & 6.958 & 7.200 & 6.797 & 8.453
& 8.243 & 6.390 & 6.189 & 9.030 & 8.207 \\
\midrule

\multirow{4}{*}{GDELT}
& 30
& 8.427 & 8.643 & \textbf{8.397} & 8.512 & 8.591 & 8.865
& 8.482 & 8.583 & \underline{8.404} & 8.480 & 9.320
& 8.443 & 8.482 & 8.967 & 9.891 & 8.492 \\

& 90
& 8.484 & 8.554 & \underline{8.413} & 8.603 & 8.614 & 8.731
& 8.515 & 8.622 & \textbf{8.314} & 9.902 & 9.741
& 8.829 & 8.576 & 8.585 & 9.450 & 8.538 \\

& 180
& 8.384 & 8.453 & \underline{8.317} & 8.420 & 8.583 & 10.197
& 8.413 & 8.742 & \textbf{8.272} & 9.317 & 9.280
& 8.444 & 8.524 & 8.656 & 9.311 & 8.386 \\

\cmidrule(l){2-18}
& \textbf{Avg}
& 8.432 & 8.550 & \underline{8.376} & 8.512 & 8.596 & 9.264
& 8.470 & 8.649 & \textbf{8.330} & 9.233 & 9.447
& 8.572 & 8.527 & 8.736 & 9.551 & 8.472 \\
\midrule

\multirow{4}{*}{Seabirds}
& 60
& 1.000 & \underline{0.916} & \textbf{0.623} & 1.466 & 2.667 & 2.951
& 1.898 & 2.011 & 2.554 & 2.839 & 2.981
& 2.005 & 3.569 & 3.800 & 2.631 & 3.111 \\

& 360
& 1.339 & \textbf{0.953} & \underline{1.168} & 1.940 & 2.827 & 1.890
& 1.634 & 2.601 & 2.253 & 1.797 & 2.191
& 2.521 & 3.025 & 3.290 & 2.851 & 3.961 \\

& 720
& 1.580 & 1.398 & \textbf{1.332} & 3.275 & 2.731 & 2.011
& \underline{1.384} & 2.551 & 2.616 & 2.220 & 2.240
& 2.380 & 2.974 & 2.001 & 2.350 & 3.030 \\

\cmidrule(l){2-18}
& \textbf{Avg}
& 1.306 & \underline{1.089} & \textbf{1.041} & 2.227 & 2.742 & 2.284
& 1.639 & 2.388 & 2.474 & 2.285 & 2.471
& 2.302 & 3.189 & 3.030 & 2.611 & 3.367 \\

\bottomrule
\end{tabular}%
}%
}

\vspace{1pt}
\parbox{\textwidth}{%
\scriptsize
\raggedright
\textit{Note.} ``-'' indicates out-of-memory (OOM) even with the batch size set to 1. Avg is computed from the three displayed prediction-length values; best and second-best markings are determined before rounding the averages.
\textit{Abbreviations.}
TFM: TFMixer;
TiW: TiWeaver;
AST: ASTGI;
KAF: KAFNet;
HiP: Hi-Patch;
HIMTS: HyperIMTS;
GNF: GNeuralFlow;
tPGNN: tPatchGNN;
GFT: GraFITi;
Warp: Warpformer;
NF: NeuralFlows.
}
\end{table}

\begin{table}[h]
\centering
\caption{Performance comparison of time series forecasting models and TSFMs in terms of MSE ($\times 10^{-2}$).
Best results are in \textbf{bold} and second-best are \underline{underlined}.}
\label{tab:regular_tsfm_mse_results}

\setlength{\tabcolsep}{2.2pt}
\renewcommand{\arraystretch}{1.08}

\makebox[\textwidth][c]{%
\resizebox{\textwidth}{!}{%
\begin{tabular}{c|c|cccccccccc}
\toprule
\textbf{Dataset} &
\textbf{Pred} &
\textbf{Crossformer} &
\textbf{DLinear} &
\textbf{PatchTST} &
\textbf{TimesNet} &
\textbf{iTransformer} &
\textbf{DuET} &
\textbf{SRSNet} &
\textbf{MIRA} &
\textbf{Sundial} &
\textbf{Timer} \\
\midrule

\multirow{4}{*}{PhysioNet}
& 1
& 1.391 & 6.965 & 6.692 & 7.573 & 6.777
& \textbf{1.115} & \underline{1.142} & 9.060 & 7.757 & 12.829 \\

& 6
& \textbf{0.850} & 5.231 & 2.538 & 2.934 & 2.431
& \underline{1.388} & 1.428 & 8.592 & 22.247 & 12.462 \\

& 12
& \textbf{0.778} & 5.261 & 2.302 & 3.105 & 2.334
& \underline{1.657} & 1.669 & 8.567 & 21.873 & 12.512 \\

\cmidrule(l){2-12}
& \textbf{Avg}
& \textbf{1.006} & 5.819 & 3.844 & 4.537 & 3.847
& \underline{1.387} & 1.413 & 8.740 & 17.292 & 12.601 \\
\midrule

\multirow{4}{*}{Human Activity}
& 200
& 0.728 & 1.647 & 0.716 & 23.004 & 1.627
& \textbf{0.265} & \underline{0.268} & 6.885 & 22.453 & 23.022 \\

& 1200
& 1.064 & 1.723 & 0.410 & 23.318 & 1.525
& \textbf{0.370} & \underline{0.380} & 10.081 & 34.614 & 23.762 \\

& 3000
& 1.987 & 1.746 & 1.218 & 23.693 & 1.544
& \textbf{0.534} & \underline{0.542} & 11.309 & 31.874 & 23.866 \\

\cmidrule(l){2-12}
& \textbf{Avg}
& 1.260 & 1.705 & 0.781 & 23.338 & 1.565
& \textbf{0.390} & \underline{0.397} & 9.425 & 29.647 & 23.550 \\
\midrule

\multirow{4}{*}{USHCN}
& 1
& 71.74 & 79.96 & 79.02 & 77.22 & 78.76
& \textbf{52.58} & \underline{58.14} & 75.51 & 84.13 & 81.35 \\

& 6
& 73.76 & 74.46 & 74.48 & 74.41 & 74.26
& \textbf{47.12} & \underline{53.21} & 72.58 & 76.90 & 74.98 \\

& 12
& 74.69 & 75.24 & 75.16 & 75.03 & 75.08
& \textbf{47.32} & \underline{53.80} & 73.45 & 77.15 & 75.88 \\

\cmidrule(l){2-12}
& \textbf{Avg}
& 73.40 & 76.55 & 76.22 & 75.55 & 76.03
& \textbf{49.01} & \underline{55.05} & 73.85 & 79.39 & 77.40 \\
\midrule

\multirow{4}{*}{CESNET}
& 24
& 2.678 & 5.196 & 3.661 & 7.349 & 5.641
& \underline{2.313} & 3.636 & \textbf{2.221} & 4.912 & 4.983 \\

& 168
& 2.563 & 5.276 & 3.417 & 7.361 & 4.765
& \underline{2.252} & 3.640 & \textbf{2.249} & 5.373 & 5.311 \\

& 336
& 2.458 & 5.524 & 3.423 & 7.422 & 5.761
& \underline{2.262} & 3.890 & \textbf{2.224} & 5.112 & 5.105 \\

\cmidrule(l){2-12}
& \textbf{Avg}
& 2.566 & 5.332 & 3.500 & 7.377 & 5.389
& \underline{2.276} & 3.722 & \textbf{2.231} & 5.132 & 5.133 \\
\midrule

\multirow{4}{*}{Pamap2}
& 200
& 0.481 & 1.150 & 2.081 & 0.784 & 2.159
& \textbf{0.277} & \underline{0.280} & 3.402 & 10.418 & 11.260 \\

& 1200
& 0.761 & 1.548 & 0.724 & 0.902 & 4.427
& \underline{0.576} & \textbf{0.572} & 6.460 & 11.244 & 12.098 \\

& 3000
& 1.073 & 1.997 & 4.487 & 1.081 & 1.994
& \underline{0.823} & \textbf{0.806} & 7.272 & 11.527 & 11.503 \\

\cmidrule(l){2-12}
& \textbf{Avg}
& 0.772 & 1.565 & 2.431 & 0.922 & 2.860
& \underline{0.559} & \textbf{0.553} & 5.711 & 11.063 & 11.620 \\
\midrule

\multirow{4}{*}{EPA-Air}
& 24
& 1.616 & 1.788 & 1.206 & 1.334 & 0.917
& \underline{0.682} & \textbf{0.628} & 6.784 & 8.385 & 7.968 \\

& 72
& 1.694 & 3.105 & 0.966 & 1.308 & 0.809
& \underline{0.698} & \textbf{0.667} & 6.818 & 8.563 & 8.351 \\

& 168
& 2.555 & 3.532 & 1.074 & 1.238 & 1.764
& \underline{0.775} & \textbf{0.720} & 7.143 & 8.514 & 8.161 \\

\cmidrule(l){2-12}
& \textbf{Avg}
& 1.955 & 2.808 & 1.082 & 1.293 & 1.163
& \underline{0.718} & \textbf{0.672} & 6.915 & 8.487 & 8.160 \\
\midrule

\multirow{4}{*}{ClusterTrace}
& 24
& 5.757 & 8.604 & 6.849 & 10.282 & 8.340
& \underline{3.996} & 4.004 & \textbf{3.964} & 10.821 & 9.629 \\

& 48
& 6.078 & 4.845 & 7.416 & 7.269 & 6.504
& \textbf{4.289} & \underline{4.340} & 4.384 & 10.333 & 10.958 \\

& 72
& 7.231 & 10.977 & 7.035 & 10.376 & 8.247
& \underline{4.623} & \textbf{4.463} & 4.664 & 9.780 & 10.528 \\

\cmidrule(l){2-12}
& \textbf{Avg}
& 6.355 & 8.142 & 7.100 & 9.309 & 7.697
& \underline{4.303} & \textbf{4.269} & 4.337 & 10.311 & 10.372 \\
\midrule

\multirow{4}{*}{APTC}
& 50
& \textbf{0.376} & \underline{0.376} & 0.377 & 0.376 & 0.376
& 0.398 & 0.400 & 0.405 & 0.379 & 0.413 \\

& 100
& 0.388 & \textbf{0.375} & \underline{0.375} & 0.375 & 0.375
& 0.383 & 0.385 & 0.415 & 0.379 & 0.396 \\

& 200
& \textbf{0.375} & \underline{0.375} & 0.378 & 0.375 & 0.375
& 0.381 & 0.385 & 0.420 & 0.378 & 0.387 \\

\cmidrule(l){2-12}
& \textbf{Avg}
& 0.380
& \textbf{0.375}
& 0.377
& \underline{0.375}
& 0.375
& 0.387
& 0.390
& 0.413
& 0.379
& 0.399 \\
\midrule

\multirow{4}{*}{FNSPID}
& 30
& 3.184 & 4.907 & 0.935 & 6.247 & 2.622
& \textbf{0.319} & 0.612 & \underline{0.422} & 6.929 & 6.716 \\

& 180
& 3.348 & 4.666 & 0.989 & 6.962 & 3.064
& \textbf{0.308} & 15.548 & \underline{0.466} & 8.500 & 8.058 \\

& 365
& 3.108 & 4.865 & 0.718 & 7.408 & 2.747
& 0.990 & \underline{0.718} & \textbf{0.430} & 7.362 & 7.097 \\

\cmidrule(l){2-12}
& \textbf{Avg}
& 3.213 & 4.813 & 0.881 & 6.872 & 2.811
& \underline{0.539} & 5.626 & \textbf{0.439} & 7.597 & 7.290 \\
\midrule

\multirow{4}{*}{GDELT}
& 30
& \textbf{1.946} & 17.667 & 4.269 & 9.949 & 8.045
& 2.034 & \underline{1.956} & 4.019 & 16.130 & 15.514 \\

& 90
& \underline{2.008} & 14.633 & 4.787 & 9.737 & 7.869
& 2.014 & \textbf{2.008} & 3.800 & 17.639 & 16.466 \\

& 180
& \textbf{1.728} & 14.461 & 5.225 & 10.076 & 7.315
& 1.859 & \underline{1.800} & 3.833 & 15.078 & 14.525 \\

\cmidrule(l){2-12}
& \textbf{Avg}
& \textbf{1.894} & 15.587 & 4.760 & 9.921 & 7.743
& 1.969 & \underline{1.921} & 3.884 & 16.282 & 15.502 \\
\midrule

\multirow{4}{*}{Seabirds}
& 60
& 2.503 & 0.179 & 0.025 & 0.034 & 0.020
& \underline{0.012} & \textbf{0.005} & 0.021 & 0.040 & 0.100 \\

& 360
& 1.974 & 0.049 & 0.158 & 0.041 & 0.028
& \textbf{0.024} & \underline{0.028} & 0.041 & 0.042 & 0.112 \\

& 720
& 3.790 & \textbf{0.023} & 0.026 & 0.050 & 0.066
& 0.045 & \underline{0.024} & 0.052 & 0.052 & 0.137 \\

\cmidrule(l){2-12}
& \textbf{Avg}
& 2.756 & 0.084 & 0.070 & 0.042 & 0.038
& \underline{0.027} & \textbf{0.019} & 0.038 & 0.045 & 0.116 \\

\bottomrule
\end{tabular}%
}%
}

\end{table}

\begin{table}[h]
\centering
\caption{Performance comparison of time series forecasting models and TSFMs in terms of MAE ($\times 10^{-2}$).
Best results are in \textbf{bold} and second-best are \underline{underlined}.}
\label{tab:regular_tsfm_mae_results}

\setlength{\tabcolsep}{2.2pt}
\renewcommand{\arraystretch}{1.08}

\makebox[\textwidth][c]{%
\resizebox{\textwidth}{!}{%
\begin{tabular}{c|c|cccccccccc}
\toprule
\textbf{Dataset} &
\textbf{Pred} &
\textbf{Crossformer} &
\textbf{DLinear} &
\textbf{PatchTST} &
\textbf{TimesNet} &
\textbf{iTransformer} &
\textbf{DuET} &
\textbf{SRSNet} &
\textbf{MIRA} &
\textbf{Sundial} &
\textbf{Timer} \\
\midrule

\multirow{4}{*}{PhysioNet}
& 1
& 9.073 & 20.636 & 20.626 & 15.178 & 19.574 & \textbf{4.020} & \underline{4.291} & 18.336 & 20.270 & 24.464 \\

& 6
& 4.845 & 17.818 & 8.244 & 8.817 & 8.688 & \textbf{4.577} & \underline{4.830} & 18.711 & 38.118 & 24.256 \\

& 12
& \textbf{4.668} & 18.031 & 9.015 & 8.948 & 8.867 & \underline{5.066} & 5.303 & 18.979 & 37.640 & 24.278 \\

\cmidrule(l){2-12}
& \textbf{Avg}
& 6.195 & 18.828 & 12.628 & 10.981 & 12.376 & \textbf{4.554} & \underline{4.808} & 18.675 & 32.009 & 24.333 \\
\midrule

\multirow{4}{*}{Human Activity}
& 200
& 5.596 & 10.224 & 5.559 & 45.665 & 10.008 & \textbf{2.992} & \underline{2.992} & 21.896 & 44.685 & 45.422 \\

& 1200
& 7.137 & 10.423 & 4.063 & 46.122 & 9.663 & \textbf{3.778} & \underline{3.779} & 28.957 & 55.996 & 46.579 \\

& 3000
& 11.066 & 10.652 & 8.091 & 46.714 & 9.722 & \textbf{4.710} & \underline{4.725} & 31.173 & 53.641 & 46.614 \\

\cmidrule(l){2-12}
& \textbf{Avg}
& 7.933 & 10.433 & 5.904 & 46.167 & 9.798 & \textbf{3.827} & \underline{3.832} & 27.342 & 51.441 & 46.205 \\
\midrule

\multirow{4}{*}{USHCN}
& 1
& 44.78 & 49.91 & 48.49 & 48.17 & 48.31 & \textbf{25.33} & \underline{33.77} & 46.48 & 57.67 & 50.54 \\

& 6
& 46.43 & 51.01 & 49.63 & 49.16 & 49.55 & \textbf{23.94} & \underline{32.28} & 47.40 & 47.04 & 49.60 \\

& 12
& 47.67 & 50.27 & 49.69 & 49.20 & 49.74 & \textbf{24.11} & \underline{32.67} & 48.90 & 47.34 & 50.11 \\

\cmidrule(l){2-12}
& \textbf{Avg}
& 46.29 & 50.40 & 49.27 & 48.84 & 49.20 & \textbf{24.46} & \underline{32.91} & 47.59 & 50.68 & 50.08 \\
\midrule

\multirow{4}{*}{CESNET}
& 24
& 11.320 & 12.912 & 11.301 & 16.692 & 13.428 & \underline{9.214} & 11.582 & \textbf{9.124} & 13.309 & 13.338 \\

& 168
& 11.210 & 13.177 & 10.954 & 16.755 & 12.480 & \textbf{9.035} & 11.521 & \underline{9.328} & 13.856 & 13.883 \\

& 336
& 10.996 & 13.340 & 10.991 & 16.810 & 14.029 & \textbf{9.071} & 11.710 & \underline{9.228} & 13.576 & 13.626 \\

\cmidrule(l){2-12}
& \textbf{Avg}
& 11.175 & 13.143 & 11.082 & 16.752 & 13.312 & \textbf{9.107} & 11.604 & \underline{9.227} & 13.580 & 13.616 \\
\midrule

\multirow{4}{*}{Pamap2}
& 200
& 5.233 & 8.522 & 10.114 & 5.493 & 10.183 & \textbf{2.864} & \underline{2.893} & 9.696 & 19.898 & 21.927 \\

& 1200
& 6.222 & 10.515 & 5.646 & 5.907 & 15.657 & \textbf{4.357} & \underline{4.372} & 14.797 & 20.700 & 22.561 \\

& 3000
& 7.826 & 11.981 & 14.849 & 6.442 & 9.674 & \textbf{5.297} & \underline{5.365} & 16.244 & 21.166 & 22.526 \\

\cmidrule(l){2-12}
& \textbf{Avg}
& 6.427 & 10.339 & 10.203 & 5.947 & 11.838 & \textbf{4.173} & \underline{4.210} & 13.579 & 20.588 & 22.338 \\
\midrule

\multirow{4}{*}{EPA-Air}
& 24
& 9.240 & 10.289 & 7.846 & 7.749 & 7.114 & \underline{5.615} & \textbf{5.149} & 14.928 & 20.295 & 19.913 \\

& 72
& 9.713 & 13.416 & 7.561 & 7.759 & 6.638 & \underline{5.538} & \textbf{5.481} & 15.358 & 20.534 & 20.430 \\

& 168
& 12.831 & 14.941 & 7.681 & 7.761 & 9.914 & \underline{6.056} & \textbf{5.813} & 16.146 & 20.471 & 20.139 \\

\cmidrule(l){2-12}
& \textbf{Avg}
& 10.595 & 12.882 & 7.696 & 7.756 & 7.889 & \underline{5.736} & \textbf{5.481} & 15.477 & 20.433 & 20.161 \\
\midrule

\multirow{4}{*}{ClusterTrace}
& 24
& 18.448 & 22.485 & 19.853 & 25.318 & 20.261 & \underline{14.963} & 15.084 & \textbf{14.901} & 25.951 & 24.494 \\

& 48
& 20.219 & 15.917 & 20.860 & 20.247 & 19.002 & \textbf{15.669} & 15.867 & \underline{15.815} & 25.559 & 26.012 \\

& 72
& 22.461 & 26.171 & 20.163 & 25.590 & 22.238 & \underline{16.166} & \textbf{15.880} & 16.220 & 24.151 & 25.021 \\

\cmidrule(l){2-12}
& \textbf{Avg}
& 20.376 & 21.524 & 20.292 & 23.718 & 20.500 & \textbf{15.599} & \underline{15.610} & 15.645 & 25.220 & 25.176 \\
\midrule

\multirow{4}{*}{APTC}
& 50
& 4.620 & 4.550 & 4.553 & \textbf{4.543} & \underline{4.543} & 4.709 & 4.728 & 4.674 & 4.566 & 4.775 \\

& 100
& 4.619 & 4.542 & 4.553 & \textbf{4.534} & \underline{4.534} & 4.616 & 4.636 & 4.727 & 4.554 & 4.669 \\

& 200
& 4.598 & 4.537 & 4.543 & \textbf{4.529} & \underline{4.529} & 4.586 & 4.612 & 4.756 & 4.548 & 4.605 \\

\cmidrule(l){2-12}
& \textbf{Avg}
& 4.612 & 4.543 & 4.550 & \textbf{4.535} & \underline{4.535} & 4.637 & 4.659 & 4.719 & 4.556 & 4.683 \\
\midrule

\multirow{4}{*}{FNSPID}
& 30
& 15.275 & 15.807 & 7.002 & 19.377 & 11.752 & \textbf{3.617} & 5.076 & \underline{3.808} & 17.798 & 18.396 \\

& 180
& 15.978 & 15.788 & 7.206 & 20.270 & 12.976 & \textbf{3.466} & 24.486 & \underline{4.570} & 20.692 & 20.765 \\

& 365
& 15.372 & 15.961 & 6.375 & 21.040 & 12.633 & 6.034 & \underline{5.228} & \textbf{4.485} & 18.126 & 18.584 \\

\cmidrule(l){2-12}
& \textbf{Avg}
& 15.542 & 15.852 & 6.861 & 20.229 & 12.454 & \underline{4.372} & 11.597 & \textbf{4.288} & 18.872 & 19.248 \\
\midrule

\multirow{4}{*}{GDELT}
& 30
& \underline{8.786} & 31.079 & 12.575 & 20.963 & 16.995 & 8.823 & \textbf{8.706} & 13.270 & 28.485 & 27.447 \\

& 90
& 9.354 & 26.966 & 12.850 & 21.124 & 17.410 & \textbf{8.759} & \underline{8.783} & 12.868 & 30.159 & 28.360 \\

& 180
& \textbf{8.440} & 26.897 & 13.021 & 21.746 & 17.266 & 8.774 & \underline{8.620} & 12.967 & 26.721 & 26.294 \\

\cmidrule(l){2-12}
& \textbf{Avg}
& 8.860 & 28.314 & 12.815 & 21.278 & 17.224 & \underline{8.785} & \textbf{8.703} & 13.035 & 28.455 & 27.367 \\
\midrule

\multirow{4}{*}{Seabirds}
& 60
& 13.227 & 3.798 & 1.075 & 1.251 & 0.988 & \underline{0.787} & \textbf{0.465} & 0.854 & 1.398 & 2.140 \\

& 360
& 12.206 & 1.774 & 2.953 & 1.411 & \underline{1.144} & \textbf{1.061} & 1.185 & 1.343 & 1.411 & 2.390 \\

& 720
& 16.814 & \textbf{1.039} & \underline{1.101} & 1.588 & 1.822 & 1.475 & 1.135 & 1.631 & 1.666 & 2.648 \\

\cmidrule(l){2-12}
& \textbf{Avg}
& 14.082 & 2.204 & 1.710 & 1.417 & 1.318 & \underline{1.108} & \textbf{0.928} & 1.276 & 1.492 & 2.393 \\
\bottomrule

\end{tabular}%
}%
}
\end{table}


\begin{figure}[h]
    \centering
    \includegraphics[width=\linewidth]{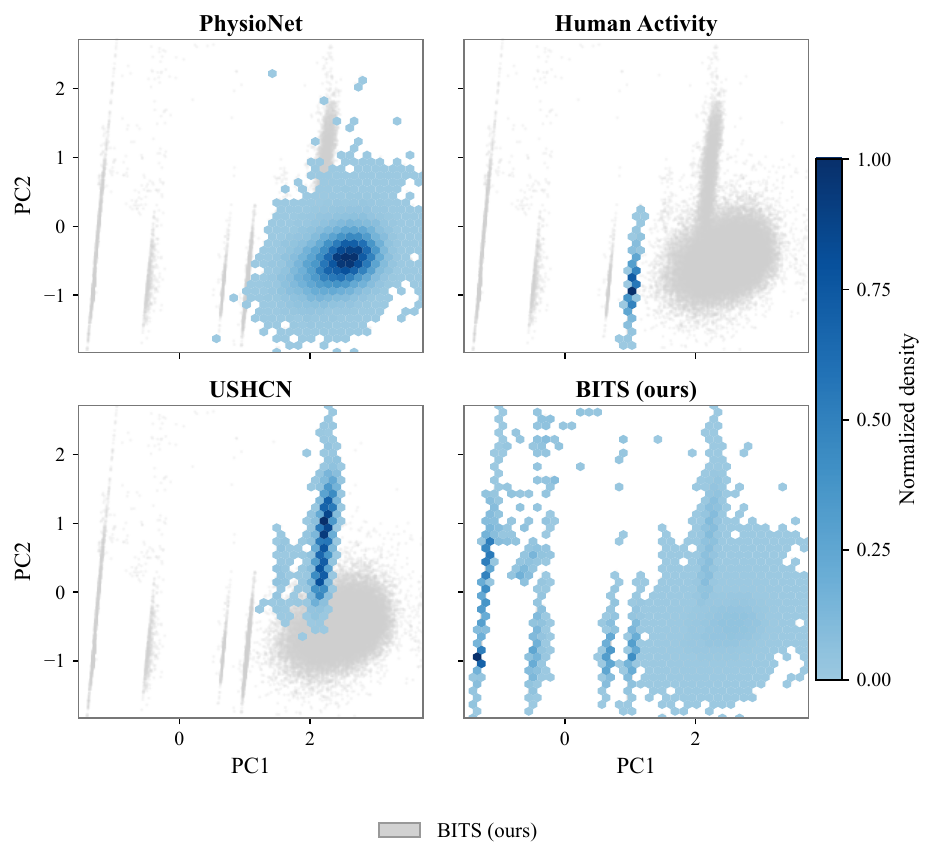}
    \caption{Coverage of irregularity characteristics across commonly used datasets and BITS. Hexbin plots visualize the distributions of four irregularity characteristics projected into a shared two-dimensional PCA space. Gray indicates the overall distribution of all 11 datasets in BITS, while blue represents the normalized density of each dataset or the entire BITS collection.
}
    \label{fig:coverage}
\end{figure}

\subsection{Leaderboard}

\begin{figure}[h]
    \centering
    \begin{subfigure}[t]{0.47\linewidth}
        \centering
        \includegraphics[width=\linewidth]{pic/mse_rank.pdf}
        \label{mse rank_appendix}
    \end{subfigure}
    \hfill
    \begin{subfigure}[t]{0.47\linewidth}
        \centering
        \includegraphics[width=\linewidth]{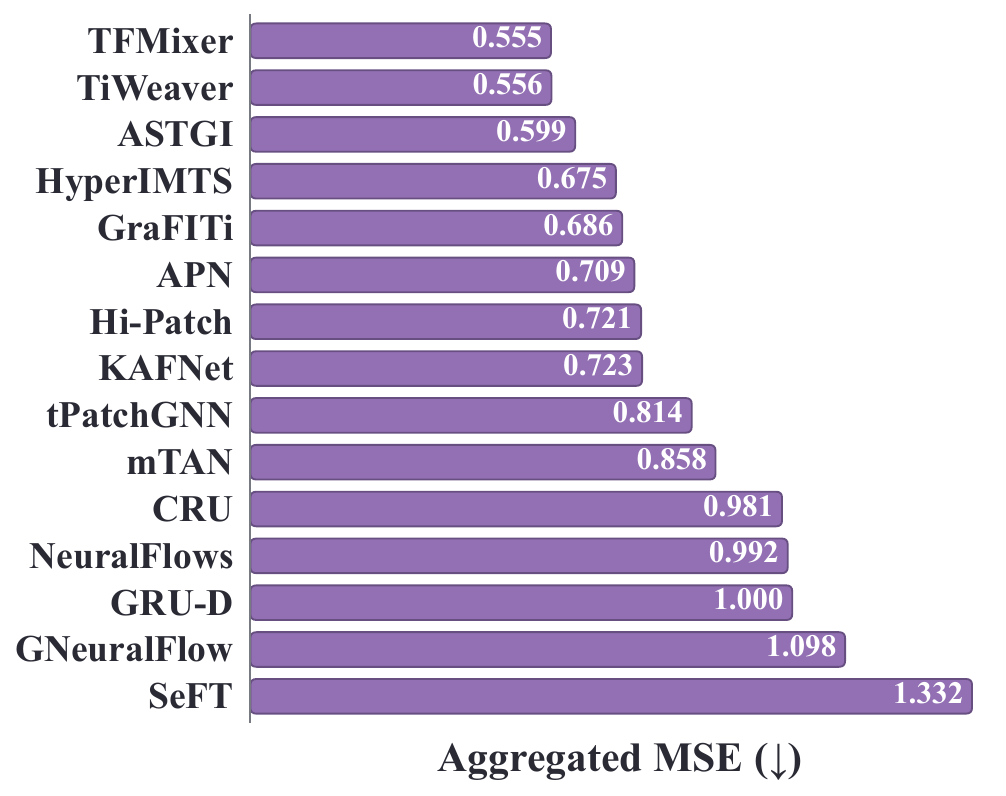}
        \label{aggregated mse rank_appendix}
    \end{subfigure}

    \centering
    \begin{subfigure}[t]{0.47\linewidth}
        \centering
        \includegraphics[width=\linewidth]{pic/dtw_rank.pdf}
        \label{dtw rank_appendix}
    \end{subfigure}
    \hfill
    \begin{subfigure}[t]{0.47\linewidth}
        \centering
        \includegraphics[width=\linewidth]{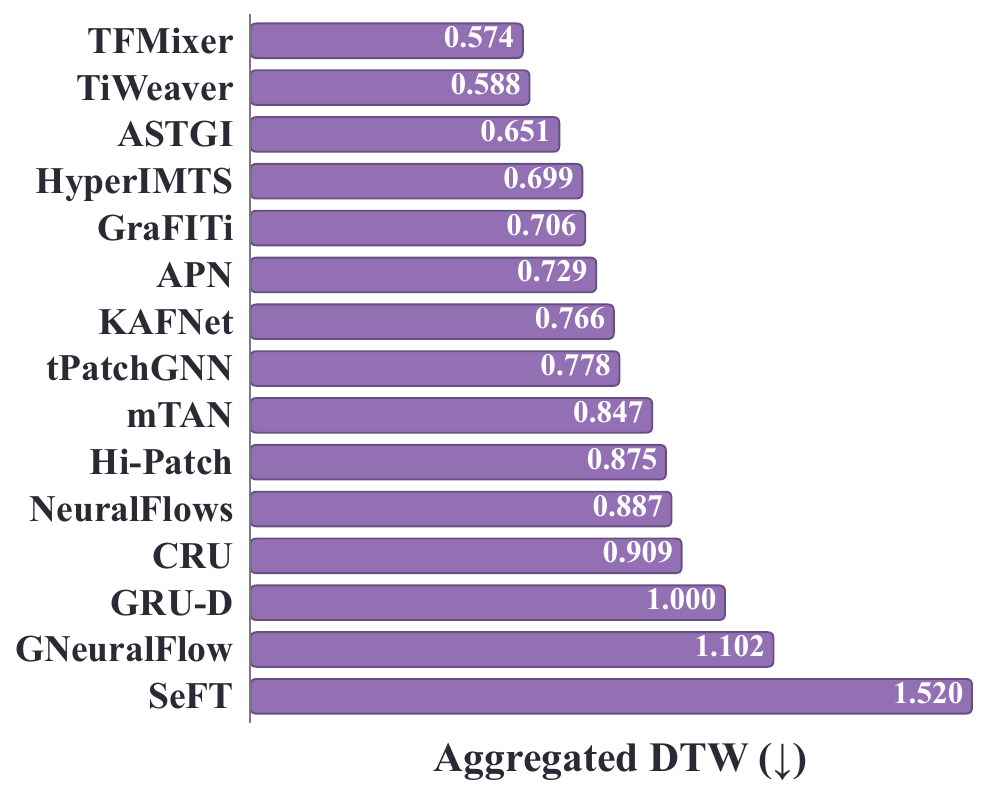}
        \label{aggregated dtw rank appendix}
    \end{subfigure}
    
    \caption{Overall performance ranking on BITS in terms of MSE rank, aggregated MSE, DTW rank and aggregated DTW. Lower is better.}
    \label{fig:rank_appendix}
\end{figure}

To provide an overall comparison, we further rank models in Figure~\ref{fig:rank_appendix}. MSE rank ranks models by MSE on each task and averages their ranks across all tasks. Aggregated MSE normalizes the MSE of each model against GRU-D on each task and computes the geometric mean of the resulting relative errors. DTW rank ranks models by DTW on each task and averages their ranks across all tasks. Aggregated DTW normalizes the DTW of each model against GRU-D on each task and computes the geometric mean of the resulting relative errors.  Lower values indicate better overall performance. Warpformer participates in per-task rankings where results are available but is excluded from the overall leaderboard due to missing results on Pamap2.

\subsection{Limitations and Future Work}
Although BITS covers diverse datasets and forecasting models, it cannot exhaustively represent all real-world irregular scenarios and emerging architectures. Future work will expand the dataset and baseline collections to accommodate new developments in irregular time series forecasting.

\end{document}